%% file: main.tex
\documentclass[10pt,twocolumn,letterpaper]{article}

\usepackage[pagenumbers]{cvpr} 

\input{preamble}

\definecolor{cvprblue}{rgb}{0.21,0.49,0.74}
\usepackage[pagebackref,breaklinks,colorlinks,allcolors=cvprblue]{hyperref}
\ifdefined\pdfsuppressptexinfo \pdfsuppressptexinfo=-1 \fi

\def\confName{CVPR}
\def\confYear{2027}

\title{CASSI-GS: Explicit Gaussian Splatting\\for Unsupervised Spectral Snapshot Compressive Imaging}

\author{Yutaro Iwamoto\\
Osaka Electro-Communication University, Osaka, Japan\\
{\tt\small yiwamoto@osakac.ac.jp}
\and
Xian-Hua Han\\
Rikkyo University, Tokyo, Japan\\
{\tt\small hanxhua@rikkyo.ac.jp}
}

\begin{document}
\maketitle
\input{sec/0_abstract}    
\input{sec/1_introduction}
\input{sec/2_related_work}
\input{sec/3_Method}
\input{sec/4_Experiments}
\input{sec/5_conclusion}
\input{sec/X_llm}
{
    \small
    \bibliographystyle{ieeenat_fullname}
    \bibliography{refs}
}

\input{sec/X_suppl_20261122.tex}

\end{document}

%% file: preamble.tex
\renewcommand{\paragraph}[1]{\vspace{.5em}\noindent\textbf{#1}}

\usepackage{bm}        
\usepackage{multirow}  
\usepackage{makecell}  
\usepackage{colortbl}  
\input{macros}         
\usepackage{multirow}
\usepackage{tabularx}  

%% file: macros.tex
\newcommand{\pix}{\mathbf{p}}                  

\newcommand{\cube}{\mathbf{X}}                 
\newcommand{\band}[1]{\mathbf{X}_{#1}}         
\newcommand{\meas}{\mathbf{Y}}                 
\newcommand{\noise}{\mathbf{E}}                
\newcommand{\mask}{\mathbf{M}}                 
\newcommand{\shiftop}[1]{\mathcal{S}_{#1}}      
\newcommand{\fwd}{\mathcal{A}}                 
\newcommand{\shiftstep}{\Delta}                
\newcommand{\nbands}{B}                        

\newcommand{\wcol}[1]{\mathbf{w}_{#1}}         

\newcommand{\nprim}{N}                         
\newcommand{\pos}[1]{\bm{\mu}_{#1}}            
\newcommand{\gaussfn}[1]{G_{#1}}               
\newcommand{\ncomp}{K}                         
\newcommand{\spec}[1]{\phi_{#1}}               
\newcommand{\amp}[1]{\alpha_{#1}}              
\newcommand{\wlc}[1]{c_{#1}}                   
\newcommand{\wlw}[1]{\sigma^{\lambda}_{#1}}    
\newcommand{\specvec}[1]{\bm{\phi}_{#1}}       
\newcommand{\shape}[1]{\hat{\bm{\phi}}_{#1}}   
\newcommand{\flux}[1]{f_{#1}}                  

\newcommand{\lpf}[1]{\rho_{#1}}                
\newcommand{\spacing}[1]{d_{#1}}               
\newcommand{\ripeps}{\varepsilon_{\mathrm{r}}} 

\newcommand{\sg}[1]{\mathrm{sg}\!\left(#1\right)}   
\newcommand{\eigA}{\lambda_1}                  
\newcommand{\eigB}{\lambda_2}                  
\newcommand{\nbrw}[1]{\omega_{#1}}             
\newcommand{\pixspec}{\mathbf{x}}              
\newcommand{\speccov}{\mathbf{C}}              
\newcommand{\speceig}[1]{\xi_{#1}}             
\newcommand{\nbrsig}{\sigma_{\mathrm{n}}}      
\newcommand{\Lcurv}{\mathcal{R}_{\mathrm{hr}}}
\newcommand{\wcurv}{w_{\mathrm{hr}}}
\newcommand{\Lnbr}{\mathcal{R}_{\mathrm{nbr}}}      
\newcommand{\Llr}{\mathcal{R}_{\mathrm{lr}}}        
\newcommand{\Ldata}{\mathcal{L}_{\mathrm{data}}}    


%% file: sec/0_abstract.tex
\begin{abstract}
Coded aperture snapshot spectral imaging (CASSI) captures a 3D spatial--spectral cube in a single compressed 2D measurement, from which the cube must be reconstructed.
Many unsupervised methods rely on deep networks, whose implicit priors are difficult to interpret.
We investigate how Gaussian splatting (GS), an explicit representation with interpretable parameters, works when applied to the severely underdetermined inverse problem of CASSI, in which the measurement alone cannot determine the placement of the primitives.
We find that (i) the standard density control of GS (splitting and spawning) does not work, so we control the density by merging only, and (ii) regularization is essential, especially against blotchy artifacts, which we suppress with a Hessian eigenvalue-ratio regularizer.
We also share 2D spatial Gaussians across all bands, assign each primitive a continuous spectrum, and suppress ripples in the sum of the Gaussians with a low-pass filter whose width is set by the primitive spacing.
The resulting method, CASSI-GS, outperforms the compared unsupervised methods on simulated data (KAIST and CAVE) and yields fewer streak and mesh-like artifacts on real data.
Moreover, unlike pixel- or network-based reconstruction, the statistics of the primitives can reveal how a regularizer works beyond quantitative scores.
\end{abstract}

%% file: sec/1_introduction.tex
\section{Introduction}
\label{sec:intro}
\begin{figure}[t]
  \centering
  \includegraphics[width=\linewidth]{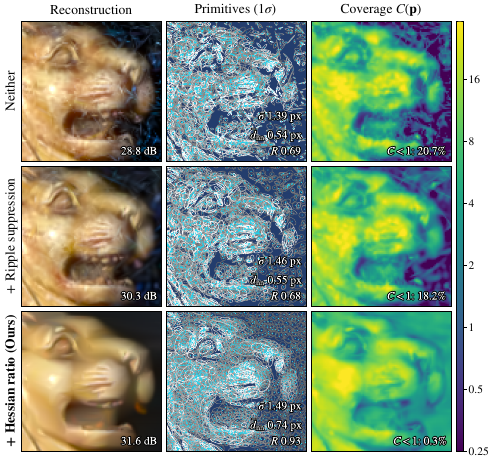}
  \caption{Effect of the ripple suppression and the Hessian eigenvalue-ratio regularizer (Hessian ratio) on the primitives (KAIST scene 10, $96 \times 96$ crop).
  Rows: neither (w/o Hessian ratio \& ripple suppression in \cref{tab:ablation}), ripple suppression only (w/o Hessian ratio), and both (ours).
  Columns: reconstruction; $1\sigma$ ellipses of the rendered width $\sqrt{\sigma^2 + \rho_i^2}$, shaded by peak intensity (dots: centers); coverage $C(\mathbf{p}) = \sum_i \tilde{G}_i(\mathbf{p})$ (log scale).
  Numbers within the crop: PSNR; median rendered width $\bar{\sigma}$, mean nearest-neighbor distance $\bar{d}_{\mathrm{nn}}$, Clark--Evans index $R$~\cite{clark1954distance} ($R = 1$: random, $R < 1$: clustered); fraction of pixels with $C < 1$.}
  \label{fig:teaser}
\end{figure}

Hyperspectral imaging (HSI) is a sensing technique that acquires spatial information together with rich spectral information and is used in a wide range of fields, including remote sensing, pharmaceutical and medical applications, and industrial inspection~\cite{hong2026hyperspectral,bhargava2024hyperspectral}.
Acquisition methods include spectral scanning, which switches a tunable filter to capture one band at a time, and push-broom scanning, which scans the scene line by line while acquiring the spatial and spectral information of each line~\cite{hong2026hyperspectral}.
However, these methods require multiple exposures and cannot capture a hyperspectral image in a single shot, which makes them unsuitable for dynamic scenes.
To address this limitation, single-disperser coded aperture snapshot spectral imaging (SD-CASSI; hereafter CASSI)~\cite{wagadarikar2008single} was proposed.
CASSI spatially modulates the 3D spatial--spectral cube with a coded aperture mask and then spectrally disperses the modulated light along one direction with a dispersive element such as a prism.
This compressive imaging technique superimposes the spectrally shifted bands onto a single 2D measurement in a single exposure, so the reconstruction is a severely underdetermined inverse problem.
Various reconstruction methods have been proposed to solve this problem accurately.

\begin{figure*}[t]
  \centering
  \includegraphics[width=\textwidth]{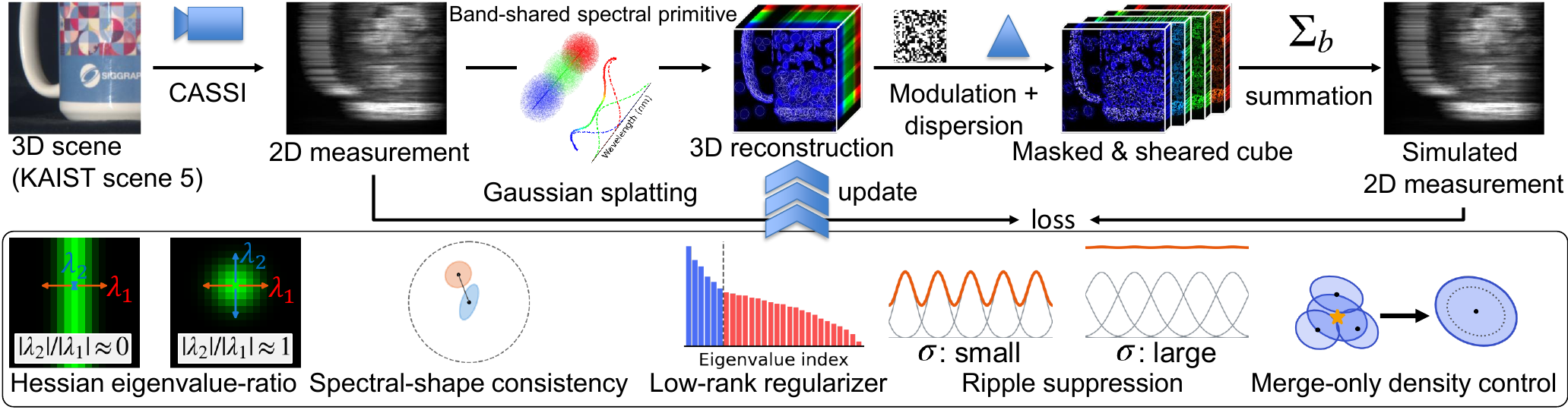}
  \caption{Schematic overview of CASSI-GS. Explicit band-shared spectral Gaussians are optimized using only a single 2D measurement.}
  \label{fig:overview}
\end{figure*}

With recent advances in deep learning, supervised methods achieve high accuracy with fast inference~\cite{zhang2024dual,zhang2024improving,huang2026lrdun}.
However, they require many pairs of simulated measurements and ground-truth cubes, which cannot be captured directly with CASSI and must be acquired with other imaging systems such as spectral scanners; they may also need retraining when the mask changes~\cite{wang2022modeling} and suffer from domain gaps across the diverse applications of HSI~\cite{zeng2024sahsci}.

Unsupervised (also called self-supervised) methods instead use the structure of an untrained network as an implicit prior and optimize it using only the CASSI measurement~\cite{meng2021selfsupervised,chen2024hyperspectral,chen2025lctcpnplctc,teng2026dbp}. Although they usually lag behind supervised methods~\cite{huang2026lrdun} in accuracy and speed, they need no training data or retraining.
The design of the network directly affects reconstruction accuracy~\cite{ulyanov2020deep}, but how combinations of layers and modules act as an implicit prior remains difficult to understand despite theoretical analyses~\cite{zhao2024untrained}, so network design often relies on trial and error.
Moreover, since the optimized quantities are the numerous network weights, it is difficult to interpret how the implicit prior affects specific structures of the reconstructed cube.
We therefore investigate how Gaussian splatting (GS)~\cite{kerbl20233d}, an explicit representation, works when applied to the severely underdetermined inverse problem of CASSI, in which the measurement alone cannot determine the placement of the primitives.
We propose CASSI-GS and address the representation, the density control, and the regularization (\cref{fig:overview}):
\begin{itemize}
\item \textbf{Explicit GS with band-shared spectral primitives.} We reconstruct the cube from a single CASSI measurement with GS (to our knowledge, the first such attempt). Because the dispersion in CASSI mixes spatial position and wavelength, each primitive has a 2D spatial Gaussian shared across all bands, which reduces the resulting ambiguity, and its own continuous spectrum. Ripples in the sum of the Gaussians are suppressed by a spacing-dependent low-pass filter.
\item \textbf{Merge-only density control.} The measurement is fitted quickly and the remaining error lies in the null space, so gradient- or error-guided splitting and spawning cannot find where to add primitives. We thus control the density only by merging nearby primitives with similar sizes and spectra.
\item \textbf{Hessian eigenvalue-ratio regularization.} To suppress blotchy artifacts, we repurpose the Hessian eigenvalue ratio used in the vesselness filter of Frangi et al.~\cite{frangi1998multiscale} as a regularizer, instead of total variation (TV) regularization~\cite{rudin1992nonlinear}. By penalizing the smaller Hessian eigenvalue relative to the larger one, it preserves lines, edges, and planar slopes while suppressing isotropic blob-like structures. The interpretable primitives show how it works: it spreads the primitives more evenly and nearly removes coverage holes (\cref{fig:teaser,sec:ablation}).

\end{itemize}
CASSI-GS outperforms the compared unsupervised methods on simulated data and yields fewer streak and mesh-like artifacts on real data.


%% file: sec/2_related_work.tex
\section{Related work}
\label{sec:related}

\paragraph{CASSI reconstruction.}
CASSI reconstruction methods can be broadly classified into model-based methods~\cite{bioucas-dias2007new,yuan2016generalized,liu2019rank}, supervised methods~\cite{meng2020endtoend,cai2022maskguided,cai2022degradationaware,zhang2024dual,zhang2024improving,huang2026lrdun}, and unsupervised methods~\cite{meng2021selfsupervised,chen2024hyperspectral,mei2024progressive,jin2025compressive,chen2025lctcpnplctc,he2026sgde}.
In supervised learning, because trained networks tend to act as black boxes, deep unfolding~\cite{cai2022degradationaware,dong2023residual,zhang2024dual,huang2026lrdun} has also been proposed; it maps each iteration of an iterative algorithm to a stage of the network, making the role of the architecture explicit.
There are also Plug-and-Play (PnP) frameworks~\cite{zheng2021deep} that use pretrained denoising networks within the optimization.

Unsupervised methods relevant to this work include methods that use the network architecture as one of their priors~\cite{meng2021selfsupervised,chen2024hyperspectral,chen2025lctcpnplctc,chen2025ldmeet,teng2026dbp} and coordinate-based continuous representations~\cite{mei2024progressive,jin2025compressive}.
All of them represent the cube, entirely or in part, as the output of a network and draw some or all of their prior from the network architecture.
In addition, since hardware-induced mask misalignment greatly degrades reconstruction on real data, SGDE~\cite{he2026sgde} jointly optimizes the reconstruction and the estimation of the misalignment from the measurement alone.

\paragraph{Gaussian splatting and inverse problems.}
3D Gaussian Splatting (3DGS)~\cite{kerbl20233d} represents a scene with 3D Gaussian primitives for high-quality, fast rendering.
2D Gaussians have also been used for image representation and compression~\cite{zhang2024gaussianimage,li2026gaussianimage} and for arbitrary-scale super-resolution~\cite{hu2025gaussiansr,chen2025generalized,peng2026pixel}.
In rendering, primitives that are small relative to the sampling interval cause aliasing and high-frequency artifacts.
Mip-Splatting~\cite{yu2024mipsplatting} and SCIGS~\cite{wang2025scigs} add the variance of a low-pass filter to the covariance of each primitive to suppress, respectively, aliasing caused by zooming in 3DGS and high-frequency artifacts in 3D reconstruction from video snapshot compressive imaging (SCI) measurements.
To represent scenes with higher quality, 3DGS~\cite{kerbl20233d} performs gradient-based density control.
Merging nearby Gaussians has also been used to reduce redundant allocation in image representation with 2D Gaussians~\cite{chen2026localityaware} and to build level-of-detail (LOD) hierarchies for large-scale scenes in 3DGS~\cite{kerbl2024hierarchical}.
In addition, GS has been applied to inverse problems in imaging, such as CT reconstruction~\cite{zha2024r2gaussian,wu2025discretized,fang20264dsg} and 3D scene recovery from video SCI (temporal compression)~\cite{wang2025scigs,yang2026gs2ci}.
In the hyperspectral domain, GS has been used for novel view synthesis from multi-view hyperspectral images~\cite{thirgood2025hypergs,narayanan2026diffusiondenoised}, hyperspectral image compression~\cite{zhang2026lrscgs,liu2025gshsi,li2025lorsgs}, and spatial arbitrary-scale super-resolution~\cite{zhang2026voronoiguided}.
For spectral CT, 4D-SG~\cite{fang20264dsg} shares Gaussian shapes across energies, but each energy is measured separately.
The work closest to ours is GSLR~\cite{zeng2026gaussian}, a GS-based low-rank tensor representation for completing missing entries of multispectral images and similar data.
It generates the latent tensor and the transform matrix of the t-SVD with 2D and 1D Gaussians, respectively, and does not address CASSI, in which all bands are superimposed in the measurement.
In \cref{sec:experiments}, we compare our method with GSLR, whose data term is replaced by that of CASSI.

\paragraph{Regularization.}
CASSI reconstruction has employed regularizers based on TV~\cite{yuan2016generalized}, nonlocal self-similarity among patches~\cite{liu2019rank,zhang2019computational}, and spectral low-rankness~\cite{chen2024hyperspectral,chen2025ldmeet}.
Image representation with 2D Gaussians also uses a regularizer that encourages nearby Gaussians to have consistent colors~\cite{chen2026localityaware}.

%% file: sec/3_Method.tex
\section{Method}
\label{sec:method}
\subsection{Preliminaries: CASSI forward model}
\label{sec:prelim}
In CASSI~\cite{wagadarikar2008single}, each band image $\band{b}\in\mathbb{R}^{H\times W}$ of the spatial--spectral cube $\cube\in\mathbb{R}^{H\times W\times\nbands}$ is spatially modulated by a coded aperture mask $\mask\in\mathbb{R}^{H\times W}$, shifted along the width direction, and summed over the bands; the resulting image is observed together with noise $\noise$:
\begin{equation}
  \meas = \fwd(\cube) + \noise, \qquad
  \fwd(\cube) = \sum_{b=1}^{\nbands} \shiftop{b}\bigl(\mask \odot \band{b}\bigr),
  \label{eq:forward}
\end{equation}
where $\shiftop{b}$ shifts an $H\times W$ image by $(b-1)\shiftstep$ pixels along the width and places it in an $H\times(W+(\nbands-1)\shiftstep)$ image, filling the rest with zeros, and $\meas\in\mathbb{R}^{H\times(W+(\nbands-1)\shiftstep)}$ is the measurement.
Recovering $\cube$ from $\meas$ is a severely underdetermined inverse problem: the unknowns far outnumber the measurements (by a factor of about 23 for the $256\times256\times28$ cubes in our simulation experiments).
Since our method fits a representation to the single measurement without training data, the null-space components are determined by the representation and the regularization.
We therefore minimize the regularizers in \cref{sec:reg} together with the data-fidelity term
\begin{equation}
  \Ldata = \frac{1}{|\meas|} \bigl\| \meas - \fwd(\cube) \bigr\|_F^2,
  \label{eq:data}
\end{equation}
where $|\mathbf{Y}|$ is the number of pixels in the measurement.
During the optimization, we also merge primitives (\cref{sec:merge}).

\subsection{Band-shared spectral primitives}
\label{sec:prim}

In 2D image compression~\cite{zhang2024gaussianimage} and super-resolution~\cite{chen2025generalized,peng2026pixel}, an image is represented as a sum of 2D Gaussians.
GaussianImage~\cite{zhang2024gaussianimage} replaces the depth sorting and alpha blending used in 3DGS with a simple summation and represents a color image as
\begin{equation}
  \begin{gathered}
    \mathbf{I}(\mathbf{p}) = \sum_{i=1}^{N} \mathbf{w}_i\, G_i(\mathbf{p}),\\
    G_i(\mathbf{p}) = \exp\Bigl(-\tfrac{1}{2}(\mathbf{p}-\boldsymbol{\mu}_i)^\top \boldsymbol{\Sigma}_i^{-1}(\mathbf{p}-\boldsymbol{\mu}_i)\Bigr),
  \end{gathered}
  \label{eq:gaussianimage}
\end{equation}
where $\boldsymbol{\mu}_i$ is the position of the $i$\nobreakdash-th Gaussian, its covariance is parameterized by a rotation $\theta_i$ and two widths $\sigma_{x,i}$ and $\sigma_{y,i}$ as $\boldsymbol{\Sigma}_i = R(\theta_i)\,\mathrm{diag}(\sigma_{x,i}^2, \sigma_{y,i}^2)\,R(\theta_i)^\top$, and $\mathbf{w}_i \in \mathbb{R}^3$ is a weighted color coefficient that merges the opacity and the RGB color; spherical harmonics are not used.
Applying this scheme directly to hyperspectral images requires as many coefficients in $\wcol{i}$ as there are bands ($\nbands=28$).
However, since spectra are generally smooth across wavelengths, independent per-band coefficients may introduce redundancy.
Moreover, in CASSI, where reconstruction is severely underdetermined, this unnecessary freedom may lead to instability.
Indeed, holding the spectrum as 28 independent per-band values instead of our three-Gaussian spectrum lowers the PSNR by 1.81\,dB in a controlled comparison (supplementary material).

In the CASSI measurement, the position along the dispersion direction is hard to separate from the wavelength.
We therefore share the spatial parameters of each primitive across all bands (as in \cite{li2025lorsgs,zhang2026lrscgs,zeng2026gaussian}), which also matches the property that spatial structures are largely shared across bands and materials are characterized by their spectral signatures.
Unlike these methods, which give each Gaussian coefficients on a global spectral basis, each of our primitives has its own spectral Gaussians, and low-rankness is not built into the representation but encouraged by a regularizer (\cref{sec:reg}).

Specifically, we replace $\wcol{i}$ in \cref{eq:gaussianimage} with a function of wavelength $\spec{i}(t)$ and represent the cube as a continuous function over the spatial--spectral domain.
To represent spectra with multiple peaks, each primitive has $\ncomp$ spectral Gaussians (the band-shared spectral primitive in \cref{fig:overview}):
\begin{equation}
  \begin{gathered}
    \cube(\pix,t) = \sum_{i=1}^{\nprim}\gaussfn{i}(\pix)\,\spec{i}(t),\\
    \spec{i}(t) = \sum_{k=1}^{\ncomp}\amp{ik}\exp\!\left(-\frac{(t-\wlc{ik})^2}{2(\wlw{ik})^2}\right),
  \end{gathered}
  \label{eq:primitive}
\end{equation}
where $t$ is a continuous wavelength coordinate in units of bands, and $\amp{ik}$, $\wlc{ik}$, and $\wlw{ik}$ are the amplitude, center, and width of the $k$-th spectral Gaussian.
The amplitude $\amp{ik}=\mathrm{sigmoid}(a_{ik})$ is kept in $(0,1)$, and the opacity is absorbed into it.
We constrain the width $\wlw{ik}$ to be at least one band so that each spectral Gaussian spans multiple bands.
The image of band $b$ is obtained by evaluating $\cube(\pix,t)$ at pixel centers $\pix$ and integer bands $t=b$, \ie, $\band{b}(\pix)=\cube(\pix,b)$.
Each spectral component of $\gaussfn{i}(\pix)\,\spec{i}(t)$ is a 3D Gaussian whose covariance is block diagonal, consisting of the $2\times2$ spatial covariance and the spectral variance.
Unlike the three spatial axes in standard 3DGS, wavelength is not a spatial axis: the position of a scene point does not change with wavelength. We therefore do not use general 3D Gaussians with rotations that couple space and wavelength.
This makes the spatial shape identical at all wavelengths, reducing the ambiguity described above, and also reduces the number of parameters (14 per primitive for $\ncomp=3$ vs.\ 33 for per-band coefficients).

\paragraph{Rendering and ripple suppression.}
For rendering, as in Mip-Splatting~\cite{yu2024mipsplatting} and SCIGS~\cite{wang2025scigs}, we add $\lpf{i}^2\mathbf{I}_2$ to the spatial covariance and rescale the peak value so that the flux is preserved:
\begin{equation}
  \begin{gathered}
    \tilde{G}_i(\mathbf{p}) = \sqrt{\frac{|\boldsymbol{\Sigma}_i|}{|\tilde{\boldsymbol{\Sigma}}_i|}}\,
      \exp\Bigl(-\tfrac{1}{2}(\mathbf{p}-\boldsymbol{\mu}_i)^\top \tilde{\boldsymbol{\Sigma}}_i^{-1}(\mathbf{p}-\boldsymbol{\mu}_i)\Bigr),\\
    \tilde{\boldsymbol{\Sigma}}_i = \boldsymbol{\Sigma}_i + \rho_i^2\, \mathbf{I}_2.
  \end{gathered}
\end{equation}
In these methods, the low-pass filter suppresses aliasing caused by Gaussians finer than the sampling interval, and its width is determined by the sampling frequency.
We additionally use it to address a different problem: ripples in the sum of neighboring Gaussians.
A sum of Gaussians of width $\sigma$ arranged on a lattice with spacing $d$ exhibits ripples with period $d$ when $\sigma$ is small relative to $d$ (see ripple suppression in \cref{fig:overview}).
By the Poisson summation formula, the relative amplitude of these ripples depends only on $\sigma/d$, and keeping it below a tolerance $\ripeps$ requires $\sigma\ge\tau d$ ($\tau\approx0.55$ for $\ripeps=0.01$).
We therefore determine the filter width of each Gaussian based on the local spacing $\spacing{i}$ between neighboring primitives:
\begin{equation}
  \rho_i = \sqrt{\max\bigl((\tau d_i)^2 - \bar{\sigma}_i^2,\ 0\bigr)} \quad \text{clipped to } [0.1, 2],
\end{equation}
where $\bar{\sigma}_i = \sqrt{\sigma_{x,i}\,\sigma_{y,i}}$ is the geometric mean of the spatial widths, $d_i$ is the local spacing estimated from the number of nearby primitives, and all quantities are in pixels. The lower bound of 0.1 pixels only slightly smooths very thin Gaussians rendered by point sampling, and the upper bound prevents excessive blurring.
For a primitive whose $\bar{\sigma}_i$ is already at least $\tau d_i$, $\rho_i$ is set to 0.1, and the ripple suppression is inactive.
The derivation is given in the supplementary material.
Both the optimization and the evaluation render the cube with $G_i$ in Eq.~(4) replaced by $\tilde{G}_i$.

\subsection{Merge-only density control}
\label{sec:merge}

3DGS~\cite{kerbl20233d} controls the density by cloning and splitting Gaussians with large positional gradients.
Image representations with 2D Gaussians also add Gaussians in regions with large errors~\cite{zhang2025imagegs,chen2026localityaware} and merge nearby Gaussians~\cite{chen2026localityaware}.
In CASSI, however, the measurement is fitted quickly, but the reconstruction error may still contain substantial null-space components that cannot be observed in the measurement, and the data term provides no gradient to correct them. Hence, splitting and spawning guided by gradients or errors cannot find where to add primitives.
Indeed, adding gradient- or error-guided splitting and spawning lowers the PSNR by 2.28 dB (\cref{tab:ablation}).
We therefore control the density by merging only, without cloning, splitting, or adding primitives.
Merging removes the redundancy of similar primitives overlapping at the same location, which makes the model lighter; moreover, how such primitives share the same signal cannot be determined from the cube, and merging assigns it to a single primitive, making the optimization better determined.
Indeed, without merging, the reconstruction accuracy drops (by 0.53 dB in PSNR; \cref{tab:ablation}).

\paragraph{Merging procedure.}
We merge primitives every 500 iterations starting from iteration 2,000.
We divide the image into grid cells and select the primitive with the largest flux in each cell as an anchor $a$.
Primitives that are close to the anchor (center distance at most $1.5\min(\bar{\sigma}_i,\bar{\sigma}_a)$), similar in size (a ratio of $\bar{\sigma}$ at most 1.5), and similar in spectral shape (cosine similarity of at least 0.99) are merged into it.
The position and covariance of the merged primitive are determined by flux-weighted moment matching~\cite{runnalls2007kullback},
\begin{equation}
  \boldsymbol{\mu}_{\mathcal{M}} = \frac{1}{F}\sum_{i\in\mathcal{M}} f_i\,\boldsymbol{\mu}_i,\quad
  \boldsymbol{\Sigma}_{\mathcal{M}} = \frac{1}{F}\sum_{i\in\mathcal{M}} f_i\bigl(\boldsymbol{\Sigma}_i + \mathbf{D}_i\bigr),
\end{equation}
and the spectral shape of the anchor is retained.
Here, $\mathcal{M}$ is the set of the anchor and the primitives merged into it, $\flux{i}=2\pi\sigma_{x,i}\sigma_{y,i}\sum_b\spec{i}(b)$ is the total flux of primitive $i$, $F=\sum_{i\in\mathcal{M}}\flux{i}$, and $\mathbf{D}_i=(\pos{i}-\pos{\mathcal{M}})(\pos{i}-\pos{\mathcal{M}})^\top$.
Merging several primitives into one lowers the local density and thins their overlap.
We therefore scale both widths by $\sqrt{2}$, regardless of the number of merged primitives, to maintain the overlap, and then rescale the amplitudes so that the total flux is preserved (as far as the amplitudes stay below 1).
The number of primitives removed in one merging step is limited to 5\% of the current number, and merging stops once the number falls below 20,000.

\subsection{Regularization}
\label{sec:reg}

Since the solution is not unique, we introduce prior knowledge to select a plausible one.
In addition to the prior knowledge already built into the Gaussian representation in \cref{sec:prim} (spatial sharing and continuous spectra), we use three regularizers that act at different scales: (i) a Hessian eigenvalue-ratio regularizer (Hessian ratio for short) that suppresses blotchy artifacts at each pixel while preserving lines and edges, (ii) a regularizer that aligns the spectral shapes of neighboring primitives, and (iii) a regularizer that encourages the spectra of the whole image to be approximated by a few bases, \ie, to be low-rank.

\paragraph{Hessian eigenvalue-ratio regularizer.}
Since a CASSI measurement is a single image in which many bands are summed, without the Hessian ratio, GS explains the measurement well but leaves blotchy artifacts in smooth regions (\cref{fig:teaser}).
We therefore need a regularizer that suppresses such blotchy artifacts and isolated blobs while preserving lines and edges.
TGV~\cite{bredies2010tgv} and Hessian Schatten norms~\cite{lefkimmiatis2013hessian} use second-order derivatives to avoid the staircase effect of TV regularization~\cite{rudin1992nonlinear}, but they penalize the magnitude of the second derivatives rather than the shape of local structures.
We therefore use as a regularizer the ratio of Hessian eigenvalues~\cite{frangi1998multiscale}, which has been used in medical image processing to distinguish line-like structures such as vessels from blob-like structures:
\begin{equation}
  \Lcurv = \frac{1}{|\Omega|}\sum_{\pix,b}
    \frac{|\eigB(\pix,b)|}{\sg{|\eigA(\pix,b)|}+\varepsilon_1\bigl(\sg{\band{b}(\pix)}+\varepsilon_2\bigr)},
  \label{eq:curv}
\end{equation}
where $\eigA$ and $\eigB$ ($|\eigA|\ge|\eigB|$) are the eigenvalues of the Hessian of the image of band $b$ at pixel $\pix$, $\band{b}(\pix)$ is the value of the reconstructed cube, $|\Omega|$ is the product of the numbers of pixels and bands, and $\mathrm{sg}(\cdot)$ is the stop-gradient operator.
On lines and edges, only one eigenvalue is large, so the penalty is nearly zero; on isolated blobs, $|\eigA|\approx|\eigB|$, so they are strongly penalized.
In flat regions, $|\eigA|$ is also small, so the denominator is dominated by $\varepsilon_1(\band{b}+\varepsilon_2)$, and variations relative to the brightness are penalized.
We divide by the brightness because, for the same relative variation, the absolute second differences are small in dark regions and the penalty would otherwise not reach them.
The constant $\varepsilon_1=0.005$ sets the boundary between flat and structured regions, and $\varepsilon_2=0.1$ prevents divergence at dark pixels.
We stop the gradients of $|\eigA|$ and $\band{b}$ in the denominator to close the loopholes of lowering the penalty by increasing the stronger curvature (\ie, sharpening edges) or by brightening pixels.

\paragraph{Neighborhood spectral-shape consistency.}
Spectra at spatially close positions are highly correlated.
Using this prior knowledge, we impose a regularizer that aligns the spectral shapes of neighboring primitives:
\begin{equation}
  \begin{gathered}
    \Lnbr = \frac{\sum_{(i,j)\in\mathcal{N}}\nbrw{ij}\,\bigl\|\shape{i}-\shape{j}\bigr\|_2^2}{\sum_{(i,j)\in\mathcal{N}}\nbrw{ij}},\\
    \nbrw{ij} = \exp\!\left(-\frac{\bigl\|\sg{\pos{i}}-\sg{\pos{j}}\bigr\|_2^2}{2\nbrsig^2}\right),
  \end{gathered}
  \label{eq:nbr}
\end{equation}
where $\specvec{i}=(\spec{i}(1),\dots,\spec{i}(\nbands))^\top$ is the vector of the spectrum of primitive $i$ (\cref{eq:primitive}) sampled at the $\nbands$ bands, and $\shape{i}$ is its spectral shape normalized to unit length.
We normalize the spectra because the brightness varies with shading even within the same object.
The neighborhood width $\nbrsig$ is set to 3 pixels, and $\mathcal{N}=\{(i,j)\mid i\neq j,\ \|\pos{i}-\pos{j}\|_2\le 3\nbrsig\}$ is the set of neighboring pairs of primitives.
We stop the gradients with respect to the positions and use the positions only to determine the neighboring pairs and their weights.
This penalty acts directly on the spectra of the primitives and changes only their spectral parameters, leaving the spatial arrangement unchanged.
We add this penalty every 10 iterations.

\paragraph{Low-rank regularizer.}
The Gaussian representation and the two regularizers above are based on local prior knowledge.
In contrast, natural hyperspectral images are known to be low-rank, \ie, the spectra of the whole image can be approximated by a few bases~\cite{chen2024hyperspectral,chen2025ldmeet}.
Using this global prior knowledge, we impose a regularizer that suppresses the spectral components outside the principal ones.
Specifically, for the covariance matrix $\speccov\in\mathbb{R}^{\nbands\times\nbands}$ of the spectra $\pixspec(\pix)$ over all pixels, we penalize the sum of its eigenvalues from the $(r+1)$-th largest onward, \ie, the smaller eigenvalues:
\begin{equation}
  \Llr = \sum_{m=r+1}^{\nbands}\speceig{m}(\speccov),
  \label{eq:lowrank}
\end{equation}
where $\pixspec(\pix)=(\band{1}(\pix),\dots,\band{\nbands}(\pix))^\top$ is the spectrum at pixel $\pix$, $\speceig{m}(\speccov)$ is the $m$-th largest eigenvalue of $\speccov$, and $r$ is the rank.
This sum corresponds to the energy that remains outside the top $r$ principal components when the spectra are represented by them.
As in LRSDN~\cite{chen2024hyperspectral}, we increase the rank $r$ during the optimization.
In our method, $r$ increases linearly from 2 to 6 with the iteration count (rounded to an integer), so that a few components determine the coarse structure early on and fine details can be represented later.

\subsection{Optimization}
\label{sec:opt}
The total loss adds the three regularizers in \cref{sec:reg} to the data term in \cref{eq:data}:
\begin{equation}
  \mathcal{L} = \mathcal{L}_{\mathrm{data}} + w_{\mathrm{hr}}\mathcal{R}_{\mathrm{hr}} + w_{\mathrm{nbr}}\mathcal{R}_{\mathrm{nbr}} + w_{\mathrm{lr}}\mathcal{R}_{\mathrm{lr}},
  \label{eq:total}
\end{equation}
with $w_{\mathrm{hr}} = 5 \times 10^{-4}$, $w_{\mathrm{nbr}} = 10^{-2}$, and $w_{\mathrm{lr}} = 3 \times 10^{-2}$.
We initialize $N = 60{,}000$ primitives on a square grid ($\sigma_{x,i} = \sigma_{y,i} = 2$ pixels, $K = 3$ spectral components with random centers) and optimize for 40,000 iterations with Adam ($\beta_1=0.9$, $\beta_2=0.999$)~\cite{kingma2015adam}, merging at regular intervals (\cref{sec:merge}).
The first 400 iterations are a warm-up without regularization that fits only the spatial parameters and amplitudes to a panchromatic coarse estimate, to place the primitives before determining their spectra.
The lower bound of the spatial widths is decreased exponentially from 2 to 0.05 pixels over the first 20,000 iterations, so that wide Gaussians first determine the coarse structure (without this coarse-to-fine schedule, the PSNR drops by 1.11 dB; \cref{tab:ablation}).
Details are in the supplementary material.

%% file: sec/4_Experiments.tex
\section{Experiments}
\label{sec:experiments}
We evaluated our method on simulated measurements synthesized by \cref{eq:forward} and on real CASSI measurements. We mainly evaluate on KAIST~\cite{choi2017highquality}, on which all hyperparameters were selected. CAVE~\cite{yasuma2010generalized} is held out: all settings are applied unchanged (see the supplementary material for the CAVE setup). For comparison, we used methods whose code or reconstruction results are publicly available. Among those that are not, DBP~\cite{teng2026dbp} reports a higher PSNR than ours on KAIST and LDMeet~\cite{chen2025ldmeet} a competitive one.
We ran the public code of LCTC~\cite{chen2025lctcpnplctc}, CEINR~\cite{mei2024progressive}, and SGDE~\cite{he2026sgde}.
For LRSDN~\cite{chen2024hyperspectral}, we used the KAIST reconstructions included in the authors' repository, and for CAVE and real data, we ported the public code to the SD-CASSI measurement model and ran it.
For GAP-TV~\cite{yuan2016generalized}, DeSCI~\cite{liu2019rank}, CST-L~\cite{cai2022coarsetofine}, and DAUHST-9stg~\cite{cai2022degradationaware}, we used the KAIST reconstructions distributed in the repository \emph{A Toolbox for Spectral Compressive Imaging}\footnote{\url{https://github.com/caiyuanhao1998/MST}} by the authors of CST and DAUHST.
Since SGDE adds the calibration of mask misalignment to LCTC, we compared with LCTC on the simulated data, which have no misalignment, and with SGDE on the real data.
GSLR~\cite{zeng2026gaussian} is the GS representation closest to ours, but its code is not publicly available; we therefore reimplemented it from the paper and applied it to the CASSI measurement model. For the methods we ran, we evaluated the result of the last of a fixed number of iterations (for LCTC, the iteration with the minimum loss, as in its public code), without using the ground truth to choose which iteration to report.

\subsection{Implementation details}
\label{sec:impl}

Our method was implemented in PyTorch~\cite{paszke2019pytorcha}, with custom CUDA kernels for the Gaussian rendering and its backward pass, and was optimized on an NVIDIA RTX A6000 GPU (48\,GB), taking about 14 min and 2.8\,GB of GPU memory per KAIST scene (a comparison with the baselines is given in the supplementary material).
The hyperparameters are given where each component is described in \cref{sec:method}, and the full list is provided in the supplementary material.
For the real data, to match the density of the simulation, we set the number of primitives to 400,000 and the lower bound for merging to 133,000.
We set $\wcurv=5\times10^{-3}$, chosen by visual inspection of the real reconstructions.
In addition, we weighted the data term by the inverse of the per-pixel noise variance (read and shot noise, with parameters estimated from the measurements) and added a learned constant offset $c$ to the forward model.
Details are given in the supplementary material.

\subsection{Evaluation on simulated data}
\label{sec:sim}

\begin{figure}[t]
  \centering
  \includegraphics[width=\linewidth]{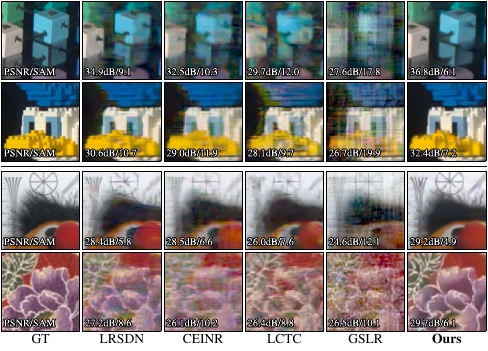}
  \caption{Visual comparison on KAIST (top two rows: scenes 2 and 8) and CAVE (bottom two rows: ``chart and stuffed toy'' and ``cloth''); $96 \times 96$ crops.
  Colors are rendered from all 28 bands with the CIE 1931 color matching functions and a common exposure for all methods.
  In each crop: PSNR (dB) / SAM ($^\circ$) of the whole scene; ours is from a single run (seed 42).}
  \label{fig:vis_sim}
\end{figure}

For the simulated data, we used the KAIST benchmark.
It consists of 10 scenes selected by TSA-Net~\cite{meng2020endtoend} from the KAIST dataset~\cite{choi2017highquality}, cropped to $256\times256$ pixels, with the spectra interpolated to 28 bands in the 450--650 nm range.
As the coded aperture mask, we used the real-valued (not binarized) mask distributed by TSA-Net~\cite{meng2020endtoend}, which is the central $256\times256$ region of a mask captured with their real system.
The measurements were simulated by \cref{eq:forward} with $\shiftstep=2$ pixels ($256\times310$ pixels), without noise.
From these measurements, we reconstructed the cubes with each method and evaluated them quantitatively (\cref{tab:qua_sim}) and visually (\cref{fig:vis_sim}).
For the quantitative evaluation, we used PSNR, SSIM~\cite{zhouwang2004image}, and SAM~\cite{kruse1993spectral} (PSNR and SSIM are computed for each band and averaged, and SAM is the angle averaged over pixels; none of them involves rounding to 8 bits).
For our method, we report the average over three runs with different random seeds.

The values reported in the literature mix results obtained with the real-valued mask, its binarized version, and a different random binary mask, and the evaluation formulas partly differ; since binary masks make reconstruction easier, we re-evaluated all methods on the same measurements with the real-valued mask and the same formulas.
Note that, as a result, some values in our tables differ from those reported in the original papers.

As shown in \cref{tab:qua_sim}, our method achieves the best PSNR, SSIM, and SAM among the compared unsupervised methods on both KAIST and CAVE, where CAVE is evaluated with the settings selected on KAIST unchanged.
Compared with the supervised methods on KAIST, our method outperforms CST-L~\cite{cai2022coarsetofine} in PSNR and SAM, and it falls behind DAUHST-9stg~\cite{cai2022degradationaware} in PSNR and SSIM.
As shown in \cref{fig:vis_sim}, our method produces sharper edges without the streaks along the dispersion direction or the grid-like patterns seen in some methods, and its colors are closer to the ground truth, consistent with its lower SAM.

\begin{table}[t]
  \centering
  \footnotesize
  \setlength{\tabcolsep}{2pt}
  \caption{Quantitative comparison on KAIST and CAVE (mean over 10 scenes each; per-scene results are in the supplementary material).
    Ours is the mean over three seeds, and the other methods are from single runs.
    Bold: best among the unsupervised methods; --: not evaluated.
    LRSDN: the authors' released reconstructions on KAIST and our port of their code on CAVE.}
  \label{tab:qua_sim}
\begin{tabular}{@{}llcccccc@{}}
\toprule
 & & \multicolumn{3}{c}{KAIST} & \multicolumn{3}{c}{CAVE} \\
\cmidrule(lr){3-5}\cmidrule(lr){6-8}
 & Method & PSNR$\uparrow$ & SSIM$\uparrow$ & SAM$\downarrow$ & PSNR$\uparrow$ & SSIM$\uparrow$ & SAM$\downarrow$ \\
\midrule
\multirow{2}{*}{\rotatebox[origin=c]{90}{\tiny Model}} & GAP-TV~\cite{yuan2016generalized} & 25.38 & 0.700 & 17.02 & -- & -- & -- \\
 & DeSCI~\cite{liu2019rank} & 27.47 & 0.781 & 12.30 & -- & -- & -- \\
\midrule
\multirow{5}{*}{\rotatebox[origin=c]{90}{\tiny Unsupervised}} & LRSDN~\cite{chen2024hyperspectral} & 36.09 & 0.938 & 5.98 & 32.61 & 0.867 & 6.73 \\
 & CEINR~\cite{mei2024progressive} & 34.60 & 0.912 & 6.88 & 32.22 & 0.843 & 7.24 \\
 & LCTC~\cite{chen2025lctcpnplctc} & 33.65 & 0.916 & 6.17 & 30.65 & 0.841 & 7.43 \\
 & GSLR~\cite{zeng2026gaussian} & 29.58 & 0.815 & 13.79 & 27.61 & 0.724 & 13.73 \\
 & \textbf{Ours} & \textbf{36.81} & \textbf{0.955} & \textbf{4.74} & \textbf{33.87} & \textbf{0.908} & \textbf{5.58} \\
\midrule
\addlinespace[3.5pt]
\multirow{2}{*}{\rotatebox[origin=c]{90}{\tiny Supervised}} & CST-L~\cite{cai2022coarsetofine} & 35.93 & 0.957 & 5.83 & -- & -- & -- \\
 & DAUHST-9stg~\cite{cai2022degradationaware} & 38.54 & 0.970 & 5.23 & -- & -- & -- \\
\addlinespace[3.5pt]
\bottomrule
\end{tabular}
\end{table}

\subsection{Evaluation on real data}
\label{sec:real}

\begin{figure}[t]
  \centering
  \includegraphics[width=\linewidth]{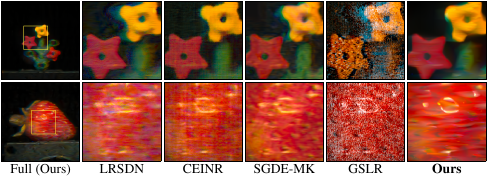}
  \caption{Results on real measurements (TSA scenes 1 and 5).
  Left: our full reconstruction with the $200 \times 200$ crop region (yellow box).
  Colors are rendered from the 28 bands as in \cref{fig:vis_sim}, but the brightness of each method is scaled by its own 99.5th percentile because the methods output different intensity scales. SGDE-MK is the multi-kernel variant of SGDE.
  Results on all five scenes are provided in the supplementary material.}
  \label{fig:real}
\end{figure}

For the real data, we used the measurements of five scenes and the coded aperture mask released by TSA-Net~\cite{meng2020endtoend}.
The measurements have $660\times714$ pixels and the mask has $660\times660$ pixels, from which we reconstruct $660\times660\times28$ cubes.
As the mask, we used the one whose misalignment was calibrated by SGDE~\cite{he2026sgde}, and shared it among all methods except SGDE, which estimates the misalignment itself.
Since no ground truth is available, we evaluated the results only visually. As shown in \cref{fig:real}, our method suppresses the streaks, mesh-like artifacts, and color speckles seen in some methods and reproduces the specular highlights on the strawberry sharply.
On the other hand, small structures such as the hole of the yellow object (scene 1) and the seeds (scene 5) are blurred in our result, and some methods reconstruct them more sharply.

\subsection{Ablation study}
\label{sec:ablation}
On the same 10 KAIST scenes as in \cref{sec:sim} (three seeds each), we removed one component at a time from the final configuration, keeping the others and the hyperparameters unchanged, and also removed both the Hessian ratio and the ripple suppression.

As shown in \cref{tab:ablation}, the full model achieves the highest PSNR, and removing the Hessian ratio causes the largest drop (3.78 dB).
Replacing the Hessian ratio with TV~\cite{rudin1992nonlinear} gives 34.83 dB, 1.98 dB below the full model. 
Comparing the middle row of \cref{fig:teaser} (ripple suppression only, \ie, w/o Hessian ratio in \cref{tab:ablation}) with the bottom row (ours), the first column shows that our reconstruction suppresses the blotchy artifacts, and the second and third columns show that the primitives become less clustered and the coverage becomes uniform, especially in the background.
Over the whole images of all scenes and seeds (30 runs), adding the Hessian ratio increases the mean nearest-neighbor distance $\bar{d}_{\mathrm{nn}}$ between primitives from 0.73 to 1.01 pixels and the Clark--Evans index $R$~\cite{clark1954distance} from 0.76 to 1.06, and decreases the fraction of pixels with coverage $C < 1$ from 14.2\% to 0.36\%, with the same direction in every run.
Removing the ripple suppression from the full model lowers the PSNR by only 0.15 dB, so its effect appears small.
However, when the Hessian ratio is also removed, the ripple suppression makes a difference of 2.03 dB.
In \cref{fig:teaser} (top vs.\ middle row), it reduces dot-like noise from thin, needle-like primitives.

\begin{table}[t]
  \centering
  \caption{Ablation study on KAIST (average over 10 scenes and three random seeds).
    w/ splitting and spawning: the full model plus splitting of the 1\% of primitives with the largest positional gradients, residual-guided spawning of up to 500 primitives, and pruning every 150 iterations up to iteration 20,000 (at most 80,000 primitives). Hessian ratio $\rightarrow$ TV: the Hessian ratio is replaced by a per-band isotropic TV.}
  \label{tab:ablation}
  \small
  \setlength{\tabcolsep}{3pt}
  \begin{tabular}{@{}lccc@{}}
    \toprule
    Configuration & PSNR$\uparrow$ & SSIM$\uparrow$ & SAM$\downarrow$ \\
    \midrule
    Full model & 36.81 & 0.955 & 4.74 \\
    \midrule
    w/o Hessian ratio & 33.03 & 0.887 & 7.11 \\
    w/o spectral-shape consistency & 35.43 & 0.935 & 5.30 \\
    w/o low-rank & 36.60 & 0.953 & 4.89 \\
    w/o merging & 36.28 & 0.949 & 5.07 \\
    w/ splitting and spawning & 34.53 & 0.939 & 4.89 \\
    w/o width scaling in merging ($\times\sqrt{2}$) & 36.52 & 0.952 & 5.18 \\
    w/o warm-up & 36.17 & 0.949 & 4.73 \\
    w/o coarse-to-fine width bound & 35.70 & 0.948 & 5.24 \\
    w/o ripple suppression & 36.66 & 0.955 & 4.74 \\
    \midrule
    w/o Hessian ratio \& ripple suppression & 31.00 & 0.829 & 8.49 \\
        Hessian ratio $\rightarrow$ TV ($w_{\mathrm{TV}}=10^{-3}$) & 34.83 & 0.927 & 6.63 \\
    \bottomrule
  \end{tabular}
\end{table}

\subsection{Application: continuous spectral rendering}
\label{sec:app}
\begin{figure}[t]
  \centering
  \includegraphics[width=\linewidth]{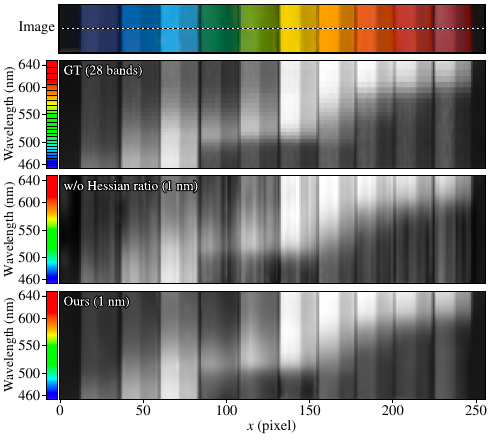}
  \caption{Continuous spectral rendering on KAIST scene 9: spatial--spectral slices along the dashed line in the top image, like a frame of a push-broom spectrometer.
  From the second row: ground truth (28 bands, each drawn over its range; ticks: band centers), and the reconstructions without the Hessian ratio and with it (ours), rendered at 1\,nm intervals. Color bars: the color of each wavelength. Spectra at individual pixels are shown in the supplementary material.}
  \label{fig:spectrum}
\end{figure}

Since each primitive holds its spectrum as a continuous function of wavelength, our representation can be rendered at arbitrary wavelengths, so the spectral sampling can be chosen freely without interpolation.
Methods based on implicit neural representations~\cite{chen2024spectralwise,chen2026phycosf} also provide continuous spectra by querying a network at each point, whereas ours renders them quickly by directly evaluating the spectral Gaussians of the primitives (\cref{eq:primitive}).
This is useful because the CASSI band centers are unequally spaced (4--11.5\,nm for KAIST) owing to the wavelength-dependent prism dispersion, whereas analyses such as spectral derivatives~\cite{tsai1998derivative} or matching with a spectral library~\cite{kokaly2017usgs} require equally spaced or specific wavelengths.
\cref{fig:spectrum} shows spatial--spectral slices.
The ground truth (second row) changes in a staircase manner around 600\,nm because each band is drawn as constant over its range, which widens at longer wavelengths, whereas our rendering at 1\,nm intervals (bottom) is smooth; the bright region moves from the lower left to the upper right, following the chalk colors (top).
Without the Hessian ratio (third row), vertical streaks run along the wavelength axis, showing spurious spatial variations within each chalk, whereas ours shows none.
Such slices provide an effective way to inspect the spatial--spectral consistency of a reconstructed cube.

\subsection{Discussion}
\paragraph{Implications.}
The interpretable parameters let us analyze how some of the regularizers work through the placement, shape, and coverage of the primitives.
Such analysis may guide the design of regularizers in other inverse problems and in GS, where regularizers have been designed separately for each task, such as sparse-view 3DGS~\cite{park2025dropgaussian,song2026d2gs,zhang2024corgs,zhu2024fsgs} and compression~\cite{lee2024compact,chen2024hac,morgenstern2024sog}, and some already use primitive-shape statistics~\cite{hyung2024effectiverank}.

\paragraph{Limitations.}
Merge-only density control requires many initial primitives and thus much memory (400,000 for the real data), and a single primitive's spectrum does not necessarily correspond to a material, which limits analyses such as spectral unmixing.

%% file: sec/5_conclusion.tex
\section{Conclusion}
\label{sec:conclusion}
We investigated how explicit GS works when applied to CASSI, a severely underdetermined inverse problem.
We found that gradient- or error-guided splitting and spawning cannot find where to add primitives and that regularization is essential against blotchy artifacts.
By combining band-shared spectral primitives, ripple suppression, merge-only density control, and the Hessian ratio, CASSI-GS achieved the highest accuracy among the compared unsupervised methods on KAIST and CAVE, and the statistics of the primitives showed how the Hessian ratio works.
Unlike pixel- or network-based reconstruction, this work demonstrates for CASSI an approach whose unknowns are explicit primitives.
Future work includes estimating the primitives with a network, representing a material with a single primitive, and extending it to other compressive inverse problems.

%% file: sec/X_llm.tex
\section*{Acknowledgements}
This work was supported in part by JSPS KAKENHI Grant-in-Aid for Scientific Research (B), Grant Number 26K02987.
\section*{LLM Usage Statement}
We used Claude (Anthropic), through the Claude Code agent, as an interactive research assistant.
In discussion with the authors, it wrote most of the research code, the evaluation scripts, the code for the baselines (porting and reimplementing them for the CASSI measurement model), the preprocessing script for an additional simulation dataset (converting the CAVE dataset to the 28-band setting), and the scripts for some figures and tables, and helped design some figures (e.g., layouts and color schemes); ran development experiments (not those reported in this paper) following the authors' instructions and summarized the results; helped search for and summarize related work; and contributed to ideation (e.g., deriving the condition on the low-pass filter width for ripple suppression from the Poisson summation formula, at the authors' request).
It also translated a draft written in the authors' native language into English and suggested revisions to the text, notation, and equations; the authors reviewed and approved all of them.
It also provided reviewer-style feedback on the manuscript.
Gemini (Google) was used as a sounding board in the early conception stage and to develop an early prototype of the code.
The authors performed the final runs reported in this paper, verified all code, reported numbers, and references, and take full responsibility for the content of this paper.

%% file: sec/X_suppl_20261122.tex
\clearpage
\setcounter{page}{1}
\maketitlesupplementary
\setcounter{section}{0}\renewcommand{\thesection}{\Alph{section}}
\setcounter{equation}{0}\renewcommand{\theequation}{S\arabic{equation}}
\setcounter{figure}{0}\renewcommand{\thefigure}{S\arabic{figure}}
\setcounter{table}{0}\renewcommand{\thetable}{S\arabic{table}}

\renewcommand{\dblfloatpagefraction}{0.5}
\makeatletter
\setlength{\@dblfptop}{0pt}\setlength{\@dblfpbot}{0pt plus 1fil}
\makeatother

This supplementary material covers the following:
\cref{sec:supp_ripple} the derivation of ripple suppression,
\cref{sec:supp_real} the forward model for real measurements,
\cref{sec:supp_wavelength} images at arbitrary wavelengths,
\cref{sec:supp_warmup} the warm-up,
\cref{sec:supp_cave} the setup for CAVE,
\cref{sec:supp_results} additional experimental results and the full list of hyperparameters, and
\cref{sec:supp_analysis} analysis of CASSI-GS.
Citation numbers refer to the references of the main paper; [S1] is listed at the end of this document.

\section{Derivation of Ripple Suppression}
\label{sec:supp_ripple}

To suppress the periodic spatial nonuniformity (ripples) that appears in the sum of neighboring primitives, we set the low-pass filter width $\rho_i$ according to the spacing between primitives (Eqs.~(5) and (6) of the main paper).
We consider the simple case of identical Gaussians of width $\sigma$ arranged with spacing $d$.

\paragraph{1D: Gaussians arranged at equal intervals.} \mbox{}\\
\textbf{Proposition 1.}
The sum $f(x)=\sum_{n\in\mathbb{Z}}g(x-nd)$ of $g(x)=\exp(-x^2/2\sigma^2)$ arranged with spacing $d$ is a continuous periodic function with period $d$ and can be written as
\begin{equation}
\begin{split}
f(x)={}&\frac{\sqrt{2\pi}\,\sigma}{d}\Bigl[1\\
&+2\sum_{k=1}^{\infty}e^{-2\pi^2k^2(\sigma/d)^2}\cos\frac{2\pi kx}{d}\Bigr].
\end{split}
\label{eq:supp_f1d}
\end{equation}

\noindent\emph{Proof.}
We use the Poisson summation formula $\sum_{n}g(x-nd)=\frac{1}{d}\sum_{k}\hat{g}(k/d)\,e^{2\pi ikx/d}$.
The Fourier transform of $g$, $\hat{g}(\xi)=\int g(x)\,e^{-2\pi i\xi x}\,dx=\sqrt{2\pi}\,\sigma\,e^{-2\pi^2\sigma^2\xi^2}$, is even, so combining the terms for $k$ and $-k$ gives \cref{eq:supp_f1d}. \hfill$\square$

The $k=0$ term $\bar{f}=\sqrt{2\pi}\,\sigma/d$ is the mean, and the oscillation around it is the ripple.
With $q=e^{-2\pi^2(\sigma/d)^2}$, the $k$-th harmonic is proportional to $q^{k^2}$ and decreases rapidly with $k^2$, so the oscillation is well approximated by the fundamental ($k=1$); for example, $q\approx2.5\times10^{-3}$ and $q^4\approx4\times10^{-11}$ for $\sigma/d=0.55$.
Hence, with $a=2q$,
\begin{equation}
f(x)\approx\bar{f}\left(1+a\cos\frac{2\pi x}{d}\right),\qquad a=2e^{-2\pi^2(\sigma/d)^2}.
\label{eq:supp_f1d_approx}
\end{equation}

\paragraph{2D: isotropic Gaussians on a square lattice.}
Since an isotropic Gaussian is separable, $G(x,y)=g(x)\,g(y)$, the sum over a square lattice with spacing $d$ is the product of the 1D sums:
\begin{equation}
\begin{split}
F(x,y)&=\sum_{m,n}g(x-md)\,g(y-nd)=f(x)\,f(y)\\
&\approx\bar{f}^{\,2}\left(1+a\cos\frac{2\pi x}{d}\right)\left(1+a\cos\frac{2\pi y}{d}\right).
\end{split}
\label{eq:supp_f2d}
\end{equation}
The maximum is $\bar{f}^{\,2}(1+a)^2$ (both cosines equal to 1, on the lattice points), and the minimum is $\bar{f}^{\,2}(1-a)^2$ (both cosines equal to $-1$, at the centers between them).
Since the mean is $\bar{F}=\bar{f}^{\,2}$, the relative ripple, defined as half the peak-to-valley difference divided by the mean, is
\begin{equation}
\begin{split}
r&:=\frac{F_{\max}-F_{\min}}{2\bar{F}}\approx\frac{(1+a)^2-(1-a)^2}{2}\\
&=2a=4\exp\!\left(-2\pi^2\frac{\sigma^2}{d^2}\right).
\end{split}
\label{eq:supp_ripple}
\end{equation}

\paragraph{Condition on the width.}
Keeping $r$ below a tolerance $\varepsilon_r$ requires
\begin{equation}
4e^{-2\pi^2(\sigma/d)^2}\le\varepsilon_r
\iff
\frac{\sigma}{d}\ge\tau:=\sqrt{\frac{\ln(4/\varepsilon_r)}{2\pi^2}}.
\label{eq:supp_tau}
\end{equation}
For $\varepsilon_r=0.01$ (1\%), $\tau\approx0.551$.

\paragraph{Filter width.}
The effective width after blurring, $\sqrt{\bar{\sigma}_i^2+\rho_i^2}$, should be at least $\tau d_i$; we therefore set
\begin{equation}
\rho_i=\sqrt{\max\left((\tau d_i)^2-\bar{\sigma}_i^2,\ 0\right)}
\label{eq:supp_rho}
\end{equation}
and clip it to $[0.1, 2]$ pixels (Eq.~(6) of the main paper), where $\bar{\sigma}_i=\sqrt{\sigma_{x,i}\sigma_{y,i}}$ is the geometric mean of the two widths of an anisotropic primitive.
The local spacing $d_i$ is estimated as follows.
We divide the image into cells of $c=4$ pixels per side and let $n_i$ be the number of primitive centers in the cell containing primitive $i$ (including $i$ itself).
Regarding the cell as a square lattice with density $n_i/c^2$, we set
\begin{equation}
d_i=\frac{c}{\sqrt{n_i}},
\label{eq:supp_spacing}
\end{equation}
\ie, the spacing when $\sqrt{n_i}$ primitives are arranged along each side.
This derivation assumes identical isotropic Gaussians on a square lattice and is an approximation for the irregular arrangements and elongated primitives encountered in practice.

\section{Forward Model for Real Measurements}
\label{sec:supp_real}

\paragraph{Simulated and real measurements.}
In the simulation, the measurement is generated as
\begin{equation}
  \meas=\fwd(\cube),\qquad \cube=\cube'+\mathbf{E}',
  \label{eq:supp_sim}
\end{equation}
where $\cube'$ is the ideal cube and $\mathbf{E}'$ is the noise in capturing the ground-truth cube, which is included in the ground truth $\cube$.
The measurement is generated from the ground truth exactly by the forward model $\fwd$, and no noise is added.
Therefore, the simulated measurement contains no errors originating from actual hardware (such as a mismatch of the forward model, an offset, or noise), and the reconstruction is evaluated against the same cube $\cube$ from which the measurement was generated (including $\mathbf{E}'$).

In contrast, a real measurement is
\begin{equation}
  \meas=\fwd'(\cube')+\mathbf{F}+\mathbf{E}'',
  \label{eq:supp_real}
\end{equation}
where $\fwd'$ is the forward operator of the actual system, which differs from the model $\fwd$ owing to, \eg, mask misalignment; $\mathbf{F}$ is a nearly constant component due to the residual black level of the detector and stray light; and $\mathbf{E}''$ is photon and read noise.
These components do not pass through $\fwd$.
However, since the problem is underdetermined, they can still be fitted by distorting the estimated cube, which produces mask-shaped patterns in the background and mesh-like patterns on objects.
We address them as follows:
\begin{enumerate}
  \item \textbf{Mask misalignment ($\fwd'\neq\fwd$):} we use the forward operator $\fwd_\theta$ with the mask calibration $\theta$ estimated by SGDE~\cite{he2026sgde} (shared with the baselines).
  \item \textbf{Constant component $\mathbf{F}$:} we model it by a learned scalar $c$ and predict the measurement as $\hat{\meas}=\fwd_\theta(\cube)+c$.
  \item \textbf{Noise $\mathbf{E}''$:} since photon noise increases with brightness, we weight the squared error by the inverse of the per-pixel noise variance.
\end{enumerate}
Accordingly, we replace Eq.~(2) of the main paper with
\begin{equation}
  \Ldata=\frac{1}{|\meas|}\sum_{p}w_p\bigl(\meas_p-\fwd_\theta(\cube)_p-c\bigr)^2,
  \label{eq:supp_data}
\end{equation}
where $w_p\propto1/\sigma_p^2$.
Following the Poissonian--Gaussian model of photon and read noise~[S1], the per-pixel variance is
\begin{equation}
    \sigma_p^2=(u\,\sigma_r)^2+\frac{u}{\eta}\max\bigl(\hat{\meas}_p,0\bigr),
  \label{eq:supp_var}
\end{equation}
where $u=\kappa/N_{\mathrm{full}}$ is 1\,DN in the units of the measurement scaled by the factor $\kappa$, chosen so that the 99.9th percentile of the coarse estimate equals 0.424 (\cref{tab:hparams_real}) ($N_{\mathrm{full}}=4{,}091$\,DN, and $3{,}208$\,DN for scene 5).
We set the read noise to $\sigma_r=5$\,DN and the conversion factor to $\eta=4$\,e$^-$/DN.
The weights are computed from the prediction at every iteration with the gradients stopped and normalized to a mean of one.

We initialize $c$ to the 1st percentile of the measurement, keep it fixed until iteration 10,000, and then update it with Adam (learning rate $10^{-4}$), keeping it non-negative.
Since $\fwd_\theta(\cube)\geq0$, the prediction cannot be lowered to $\meas$ at pixels where $\meas<c$, and the difference remains as a residual.
As a result, the reconstruction becomes zero in the background instead of filling it with patterns.

We use $c$ and the noise weighting only in our method and keep the original forward models for the baselines.
Since both are improvements of the measurement model, they can be used regardless of the reconstruction method.

\section{Images at Arbitrary Wavelengths}
\label{sec:supp_wavelength}
In Eq.~(4) of the main paper, $t$ is a continuous wavelength coordinate in units of bands, and the integer $t=b$ corresponds to band $b$ ($b=1,\dots,28$).
An arbitrary wavelength $\lambda$ ($453.5\leq\lambda\leq648.0$\,nm) is converted to $t$ by linear interpolation between the center wavelengths of adjacent bands:
\begin{equation}
  t(\lambda)=b+\frac{\lambda-\lambda_b}{\lambda_{b+1}-\lambda_b},
  \qquad \lambda_b\leq\lambda\leq\lambda_{b+1},
  \label{eq:supp_wavelength}
\end{equation}
where $\lambda_b$ is the center wavelength of band $b$ and $b=1,\dots,27$.
Substituting $t(\lambda)$ into Eq.~(4) of the main paper gives the image at wavelength $\lambda$, $\cube(\pix,t(\lambda))$.
The images at 1\,nm intervals in Fig.~5 of the main paper were rendered in this way.

\section{Warm-up}
\label{sec:supp_warmup}
In the first 400 iterations, we fix the centers $\wlc{ik}$ and widths $\wlw{ik}$ of the spectral Gaussians and optimize only the spatial parameters (positions, widths, and rotations) and the amplitudes $\amp{ik}$.
During this stage, we use neither the data term on the measurement nor the regularizers, and fit the primitives to a panchromatic image computed from the measurement.
The panchromatic image is computed as follows.
First, we sum the mask $\mask$ over all bands while shifting it by $\shiftstep$ pixels per band to obtain $\Phi\in\mathbb{R}^{H\times(W+(\nbands-1)\shiftstep)}$, which represents the total transmittance of the light reaching each pixel of the measurement, and divide the measurement $\meas$ by $\Phi$ to normalize the amount of light.
Next, we crop $H\times W$ regions from the result while shifting by $\shiftstep$ pixels, stack them as the band images, and sum them along the wavelength.
We initialize the primitives by fitting them to this coarse estimate.

\section{Setup for CAVE}
\label{sec:supp_cave}

\begin{figure}[t]
  \centering
  \includegraphics{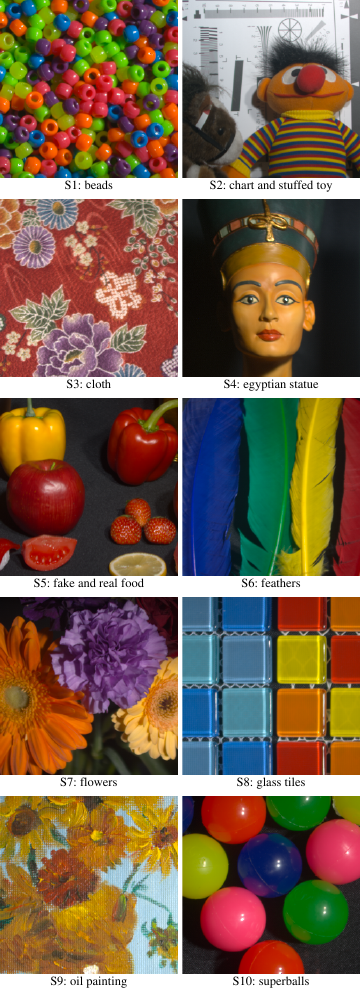}
  \caption{The ten CAVE scenes used in our experiments ($256\times256$ crops). Colors are rendered from the 28 bands with the CIE 1931 color matching functions.}
  \label{fig:supp_cave_scenes}
\end{figure}

For the simulation experiments on CAVE~\cite{yasuma2010generalized}, we used ten scenes: beads, chart and stuffed toy, cloth, egyptian statue, fake and real food, feathers, flowers, glass tiles, oil painting, and superballs.
Each of these scenes has 31 bands from 400 to 700 nm at 10 nm intervals, with 512$\times$512 pixels and 16 bits.
Without looking at any reconstruction results, we selected the ten scenes only from the RGB images provided with CAVE so that they include diverse materials and texture patterns, and cropped a $256\times256$ region from each scene without downsampling (\cref{fig:supp_cave_scenes}).
The values were divided by 65535, without per-scene normalization.
To match the KAIST benchmark built by TSA-Net~\cite{meng2020endtoend}, the 31 bands of each pixel were interpolated with a cubic spline to the same 28 center wavelengths (453.5--648.0\,nm), and negative values caused by the interpolation were set to zero.
For the CASSI forward model, as in the KAIST experiments, we used the real-valued mask distributed by TSA-Net~\cite{meng2020endtoend} with a dispersion shift of $\shiftstep=2$ pixels per band.

\section{Additional Experimental Results}
\label{sec:supp_results}
\begin{figure*}[p]
  \centering
  \includegraphics[width=\textwidth]{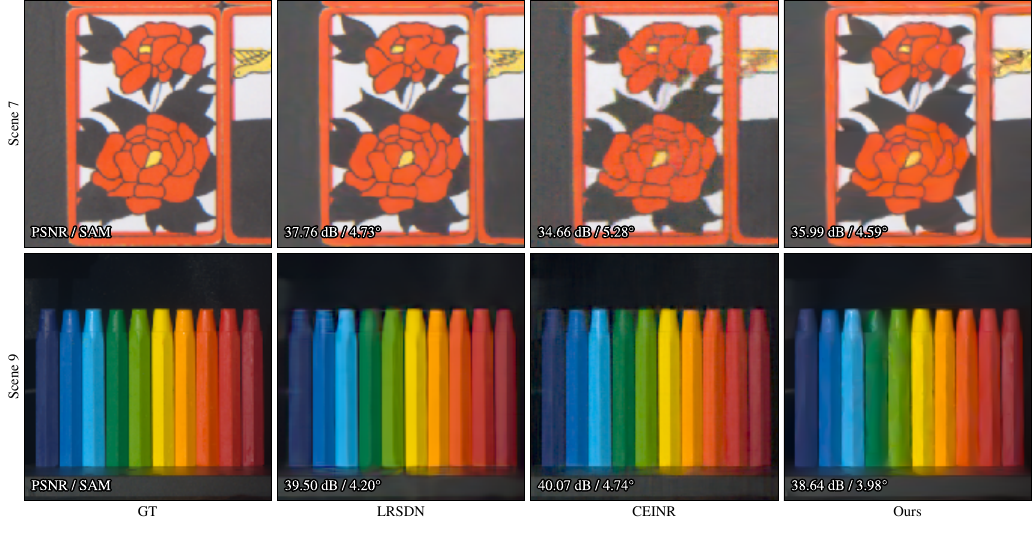}
  \caption{Reconstructions of KAIST scenes 7 (top) and 9 (bottom). Colors are rendered from the 28 bands with the CIE 1931 color matching functions as in~\cref{fig:vis_sim} of the main paper, with a common exposure for each scene. Ours is a single run with random seed 42, and LRSDN is the reconstruction released by its authors. Numbers: PSNR (dB) / SAM ($^\circ$) of the whole scene for these runs; Tab.~\ref{tab:supp_kaist_scenes} reports the mean over three seeds for ours.}
  \label{fig:supp_kaist_s7_s9}
  \vspace{1.5em}
  \includegraphics[width=\textwidth]{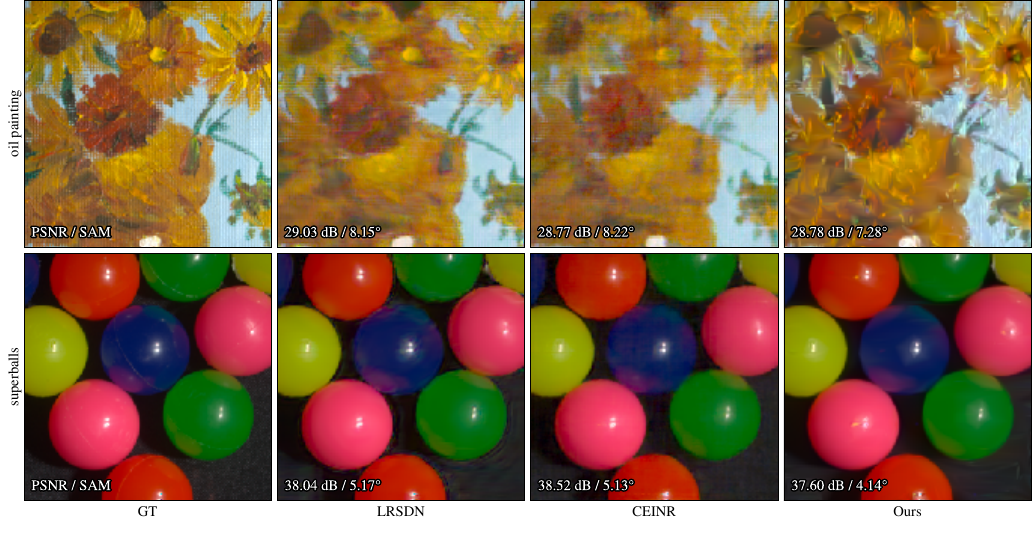}
  \caption{Reconstructions of two CAVE scenes (top: oil painting, S9; bottom: superballs, S10 in Tab.~\ref{tab:supp_cave_scenes}). Colors are rendered from the 28 bands with the CIE 1931 color matching functions as in Fig.~3 of the main paper, with a common exposure for each scene. Ours is a single run with random seed 42, and LRSDN is our port of the authors' code. Numbers: PSNR (dB) / SAM ($^\circ$) of the whole scene for these runs; Tab.~\ref{tab:supp_cave_scenes} reports the mean over three seeds for ours.}
  \label{fig:supp_cave_s9_s10}
\end{figure*}

Tables~\ref{tab:supp_kaist_scenes} and~\ref{tab:supp_cave_scenes} list the per-scene results on KAIST and CAVE, respectively. Tables~\ref{tab:hparams} and \ref{tab:hparams_real} list all hyperparameters. Among the unsupervised methods, ours achieves the highest PSNR in eight of the ten scenes on each dataset. The exceptions are KAIST scenes 7 and 9 and the CAVE scenes oil painting and superballs, where the following methods achieve a higher PSNR: LRSDN~\cite{chen2024hyperspectral} in scene 7; LRSDN, CEINR~\cite{mei2024progressive}, and LCTC~\cite{chen2025lctcpnplctc} in scene 9; LRSDN in oil painting; and LRSDN and CEINR in superballs. In three of these four scenes (KAIST scene 9, oil painting, and superballs), our method nevertheless achieves the lowest SAM. 
Figures~\ref{fig:supp_kaist_s7_s9} and~\ref{fig:supp_cave_s9_s10} show the reconstructions of KAIST scenes 7 and 9 and of the CAVE scenes oil painting and superballs, respectively. In all of these scenes, our PSNR is lower than that of at least one compared method. In KAIST scene 7, the reconstruction of CEINR is blurred, and ours shows few blotches overall but loses fine details in the yellow region at the upper right. In scene 9, our method smooths the boundaries between the faces of the chalk sticks, so that their shapes appear blurred, especially for the green ones. In oil painting, LRSDN and CEINR are blurred overall, whereas ours has sharper edges but shows elongated, brushstroke-like structures. In superballs, LRSDN shows wavy contours outside the edges of the balls. Ours reproduces the edges of the balls cleanly, but adds spurious highlights in a few places.

\begin{figure*}[p]
  \centering
  \includegraphics[width=\textwidth]{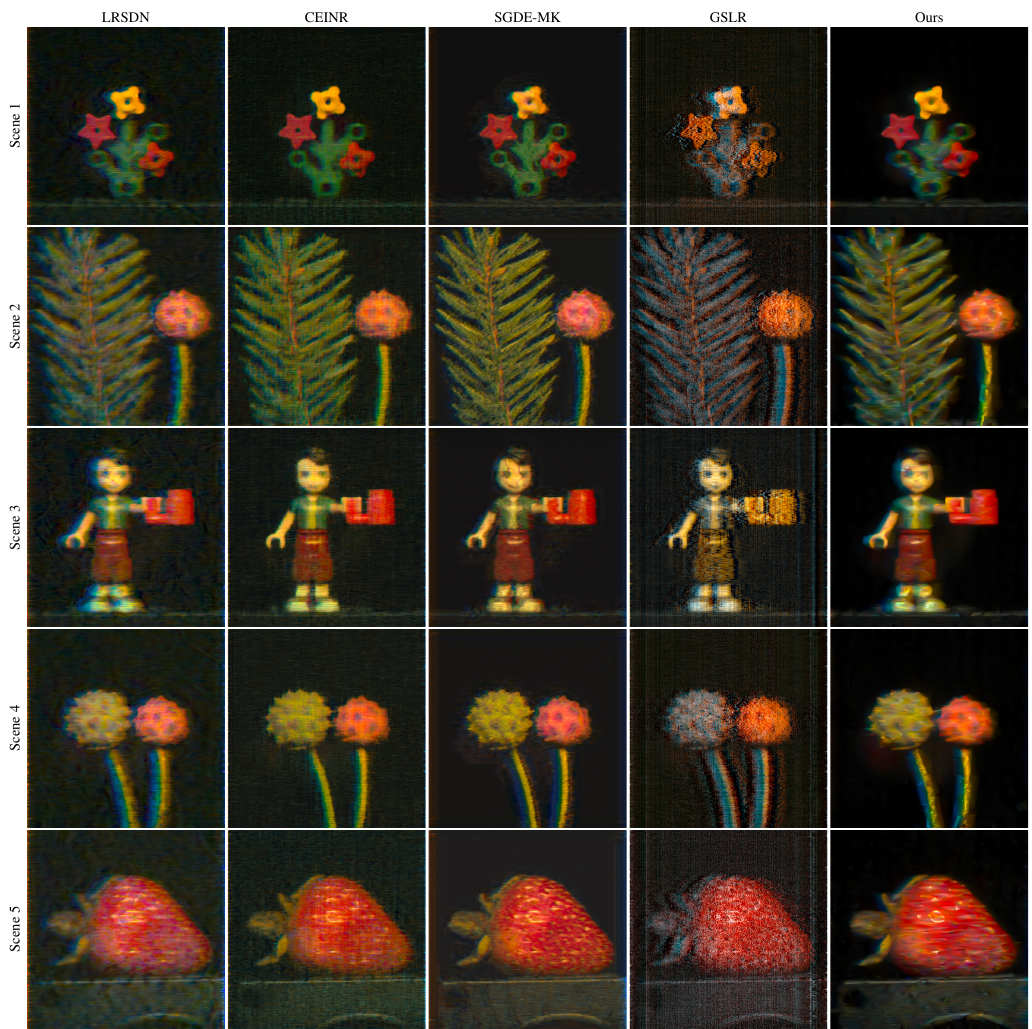}
  \caption{Reconstructions of all five real scenes (TSA-Net). Colors are rendered from the 28 bands with the CIE 1931 color matching functions as in Fig.~3 of the main paper, but the brightness of each method is scaled by its own 99.5th percentile because the methods output different intensity scales. SGDE-MK is the multi-kernel variant of SGDE~\cite{he2026sgde}.}
  \label{fig:supp_real_all}
\end{figure*}

Figure~\ref{fig:supp_real_all} shows the reconstructions of all five real scenes. Overall, CEINR and our method give visually favorable results. In scene 2, CEINR shows mesh-like artifacts but reproduces the stem at the lower right cleanly, whereas ours has no mesh-like artifacts but blurs the stem horizontally.



\begin{table*}[p]
\centering
\footnotesize
\setlength{\tabcolsep}{3pt}
\caption{Per-scene results on KAIST (PSNR in dB, SSIM, and SAM in degrees). Ours is the mean over three seeds, and the other methods are from single runs. Bold: best among the unsupervised methods. Avg: mean over the 10 scenes (Tab.~1 of the main paper).}
\label{tab:supp_kaist_scenes}
\begin{tabular}{@{}llccccccccccc@{}}
\toprule
Method & Metric & S1 & S2 & S3 & S4 & S5 & S6 & S7 & S8 & S9 & S10 & Avg \\
\midrule
GAP-TV~\cite{yuan2016generalized} & PSNR & 25.93 & 24.73 & 25.80 & 36.43 & 23.19 & 22.45 & 23.45 & 22.63 & 24.87 & 24.33 & 25.38 \\
 & SSIM & 0.723 & 0.637 & 0.778 & 0.892 & 0.689 & 0.661 & 0.666 & 0.656 & 0.720 & 0.580 & 0.700 \\
 & SAM & 13.73 & 17.79 & 14.07 & 13.60 & 14.49 & 20.90 & 16.07 & 24.04 & 14.08 & 21.40 & 17.02 \\
\cmidrule(l){2-13}
DeSCI~\cite{liu2019rank} & PSNR & 27.95 & 26.32 & 29.10 & 40.07 & 25.53 & 24.33 & 25.49 & 23.60 & 27.43 & 24.89 & 27.47 \\
 & SSIM & 0.787 & 0.690 & 0.876 & 0.964 & 0.775 & 0.750 & 0.764 & 0.739 & 0.810 & 0.652 & 0.781 \\
 & SAM & 11.01 & 15.69 & 8.22 & 6.41 & 11.29 & 15.41 & 11.23 & 19.14 & 9.51 & 15.14 & 12.30 \\
\midrule
LRSDN~\cite{chen2024hyperspectral} & PSNR & 35.44 & 34.89 & 38.90 & 45.29 & 34.71 & 33.19 & \textbf{37.76} & 30.57 & 39.50 & 30.62 & 36.09 \\
 & SSIM & 0.923 & 0.910 & 0.961 & 0.985 & 0.950 & 0.931 & \textbf{0.964} & 0.902 & \textbf{0.963} & 0.889 & 0.938 \\
 & SAM & 5.62 & 9.13 & 4.08 & 3.56 & 3.54 & 7.58 & 4.73 & 10.70 & 4.20 & 6.67 & 5.98 \\
\cmidrule(l){2-13}
CEINR~\cite{mei2024progressive} & PSNR & 34.18 & 32.51 & 38.33 & 45.98 & 32.65 & 30.75 & 34.66 & 28.96 & \textbf{40.07} & 27.94 & 34.60 \\
 & SSIM & 0.902 & 0.871 & 0.958 & \textbf{0.988} & 0.930 & 0.894 & 0.939 & 0.874 & 0.962 & 0.802 & 0.912 \\
 & SAM & 6.85 & 10.27 & 4.14 & \textbf{3.17} & 5.17 & 8.99 & 5.28 & 11.88 & 4.74 & 8.28 & 6.88 \\
\cmidrule(l){2-13}
LCTC~\cite{chen2025lctcpnplctc} & PSNR & 33.88 & 29.69 & 32.99 & 44.95 & 31.83 & 31.16 & 34.84 & 28.08 & 39.23 & 29.86 & 33.65 \\
 & SSIM & 0.914 & 0.825 & 0.934 & 0.985 & 0.929 & 0.913 & 0.958 & 0.862 & 0.962 & 0.873 & 0.916 \\
 & SAM & 6.17 & 11.97 & 5.10 & 3.77 & 4.15 & 6.64 & \textbf{4.36} & 9.74 & 4.67 & 5.10 & 6.17 \\
\cmidrule(l){2-13}
GSLR~\cite{zeng2026gaussian} & PSNR & 31.37 & 27.63 & 28.51 & 41.73 & 27.84 & 28.54 & 27.37 & 26.70 & 29.05 & 27.10 & 29.58 \\
 & SSIM & 0.849 & 0.759 & 0.800 & 0.953 & 0.824 & 0.800 & 0.743 & 0.793 & 0.865 & 0.768 & 0.815 \\
 & SAM & 10.04 & 17.80 & 12.85 & 10.70 & 8.93 & 17.97 & 11.04 & 19.88 & 11.96 & 16.71 & 13.79 \\
\cmidrule(l){2-13}
Ours & PSNR & \textbf{36.13} & \textbf{36.79} & \textbf{39.66} & \textbf{46.80} & \textbf{35.44} & \textbf{34.34} & 35.93 & \textbf{32.35} & 38.57 & \textbf{32.07} & \textbf{36.81} \\
 & SSIM & \textbf{0.941} & \textbf{0.947} & \textbf{0.969} & 0.988 & \textbf{0.966} & \textbf{0.953} & 0.957 & \textbf{0.938} & 0.962 & \textbf{0.928} & \textbf{0.955} \\
 & SAM & \textbf{4.79} & \textbf{6.14} & \textbf{3.16} & 4.01 & \textbf{2.95} & \textbf{5.63} & 4.60 & \textbf{7.11} & \textbf{4.00} & \textbf{4.98} & \textbf{4.74} \\
\midrule
CST-L~\cite{cai2022coarsetofine} & PSNR & 35.84 & 36.57 & 37.56 & 42.48 & 33.43 & 35.70 & 34.45 & 33.80 & 36.11 & 33.41 & 35.93 \\
 & SSIM & 0.948 & 0.955 & 0.962 & 0.978 & 0.955 & 0.965 & 0.938 & 0.963 & 0.954 & 0.952 & 0.957 \\
 & SAM & 6.61 & 6.76 & 4.71 & 5.78 & 4.45 & 5.95 & 5.64 & 6.83 & 5.69 & 5.89 & 5.83 \\
\cmidrule(l){2-13}
DAUHST-9stg~\cite{cai2022degradationaware} & PSNR & 37.31 & 39.15 & 41.27 & 47.01 & 35.88 & 37.21 & 37.64 & 35.13 & 40.14 & 34.62 & 38.54 \\
 & SSIM & 0.960 & 0.969 & 0.974 & 0.988 & 0.971 & 0.973 & 0.965 & 0.969 & 0.972 & 0.960 & 0.970 \\
 & SAM & 5.20 & 6.04 & 3.45 & 5.57 & 3.62 & 6.09 & 4.20 & 7.58 & 4.37 & 6.21 & 5.23 \\
\bottomrule
\end{tabular}
\end{table*}

\begin{table*}[p]
\centering
\footnotesize
\setlength{\tabcolsep}{3pt}
\caption{Per-scene results on CAVE (PSNR in dB, SSIM, and SAM in degrees). Scenes: S1 beads, S2 chart and stuffed toy, S3 cloth, S4 egyptian statue, S5 fake and real food, S6 feathers, S7 flowers, S8 glass tiles, S9 oil painting, S10 superballs. Ours is the mean over three seeds, and the other methods are from single runs. Bold: best among the unsupervised methods. Avg: mean over the 10 scenes (Tab.~1 of the main paper). LRSDN: our port of the authors' code.}
\label{tab:supp_cave_scenes}
\begin{tabular}{@{}llccccccccccc@{}}
\toprule
Method & Metric & S1 & S2 & S3 & S4 & S5 & S6 & S7 & S8 & S9 & S10 & Avg \\
\midrule
LRSDN~\cite{chen2024hyperspectral} & PSNR & 24.85 & 28.36 & 27.17 & 40.70 & 36.85 & 37.80 & 32.84 & 30.51 & \textbf{29.03} & 38.04 & 32.61 \\
 & SSIM & 0.748 & 0.867 & 0.720 & 0.978 & 0.942 & 0.948 & 0.860 & 0.927 & 0.715 & 0.967 & 0.867 \\
 & SAM & 10.27 & 5.83 & 8.64 & \textbf{6.71} & 5.67 & 4.32 & 8.94 & 3.61 & 8.15 & 5.17 & 6.73 \\
\cmidrule(l){2-13}
CEINR~\cite{mei2024progressive} & PSNR & 23.77 & 28.50 & 26.13 & 40.18 & 36.22 & 38.60 & 31.95 & 29.57 & 28.77 & \textbf{38.52} & 32.22 \\
 & SSIM & 0.664 & 0.852 & 0.634 & 0.971 & 0.938 & 0.954 & 0.836 & 0.903 & 0.711 & 0.968 & 0.843 \\
 & SAM & 12.14 & 6.56 & 10.21 & 7.46 & 5.55 & 3.88 & 9.40 & 3.82 & 8.22 & 5.13 & 7.24 \\
\cmidrule(l){2-13}
LCTC~\cite{chen2025lctcpnplctc} & PSNR & 22.62 & 26.02 & 26.45 & 38.37 & 34.71 & 37.17 & 32.22 & 29.52 & 28.29 & 31.12 & 30.65 \\
 & SSIM & 0.657 & 0.791 & 0.705 & 0.957 & 0.925 & 0.948 & 0.856 & 0.908 & 0.723 & 0.936 & 0.841 \\
 & SAM & 12.55 & 7.59 & 8.77 & 9.07 & 5.62 & 4.21 & 8.22 & 3.75 & 7.98 & 6.53 & 7.43 \\
\cmidrule(l){2-13}
GSLR~\cite{zeng2026gaussian} & PSNR & 17.43 & 24.58 & 26.53 & 39.01 & 30.97 & 26.93 & 28.35 & 25.87 & 27.61 & 28.84 & 27.61 \\
 & SSIM & 0.332 & 0.673 & 0.694 & 0.944 & 0.816 & 0.751 & 0.750 & 0.802 & 0.727 & 0.755 & 0.724 \\
 & SAM & 27.07 & 12.13 & 10.13 & 11.10 & 14.05 & 12.68 & 14.35 & 9.48 & 10.39 & 15.88 & 13.73 \\
\cmidrule(l){2-13}
Ours & PSNR & \textbf{26.64} & \textbf{29.30} & \textbf{29.70} & \textbf{43.07} & \textbf{37.14} & \textbf{39.63} & \textbf{35.33} & \textbf{31.71} & 28.85 & 37.30 & \textbf{33.87} \\
 & SSIM & \textbf{0.833} & \textbf{0.915} & \textbf{0.847} & \textbf{0.982} & \textbf{0.950} & \textbf{0.966} & \textbf{0.919} & \textbf{0.941} & \textbf{0.756} & \textbf{0.973} & \textbf{0.908} \\
 & SAM & \textbf{8.11} & \textbf{4.86} & \textbf{6.08} & 8.64 & \textbf{4.56} & \textbf{3.25} & \textbf{5.51} & \textbf{3.43} & \textbf{7.22} & \textbf{4.10} & \textbf{5.58} \\
\bottomrule
\end{tabular}
\end{table*}


\begin{figure*}[p]
  \centering
  \includegraphics[width=\textwidth]{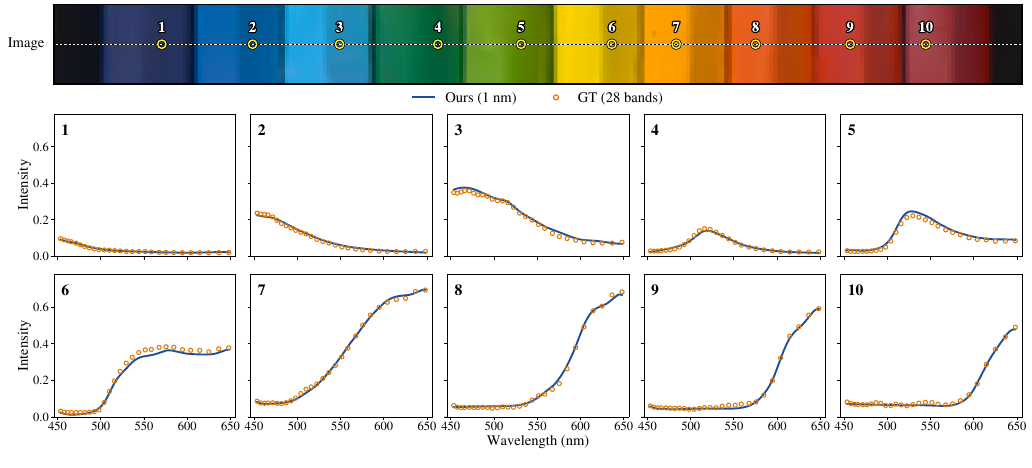}
  \caption{Spectra at individual pixels on KAIST scene 9 (the same row as Fig.~5 of the main paper).
  Top: our reconstruction (random seed 42) around the row (dashed line) with the ten pixels marked and numbered.
  Bottom: spectrum at each pixel (lines: ours rendered at 1\,nm intervals; circles: ground truth at the 28 bands).
  No ground truth is available between the bands, so accuracy there is not evaluated.}
  \label{fig:supp_spectra}
\end{figure*}

\begin{figure*}[p]
  \centering
  \includegraphics[width=\textwidth]{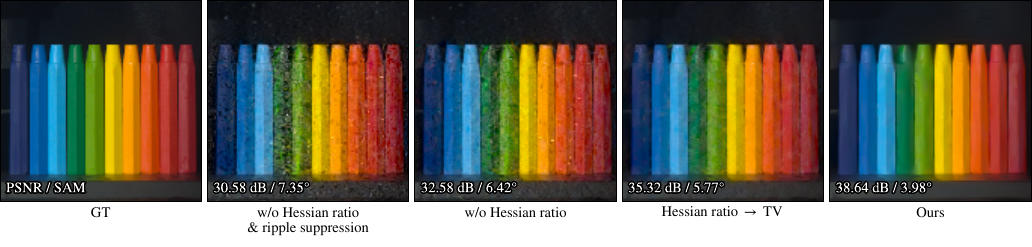}
\caption{Blotchy artifacts on KAIST scene 9. From left: ground truth, without the Hessian ratio and ripple suppression, without the Hessian ratio, the Hessian ratio replaced by TV ($w_{\mathrm{TV}}=10^{-3}$), and ours (Tab.~2 of the main paper). All reconstructions are from a single run with random seed 42, and colors are rendered as in Fig.~3 of the main paper with a common exposure. Numbers: PSNR (dB) / SAM ($^\circ$) of the whole scene for this run; Tab.~\ref{tab:supp_kaist_scenes} reports the mean over three seeds for ours.}
  \label{fig:supp_blotch}
\end{figure*}

\begin{table}[t]
\centering
\caption{Comparison of primitive representations on KAIST (mean over 10 scenes and three seeds). All three are without the neighborhood spectral-shape consistency and the merging, with 60,000 primitives.}
\label{tab:repr}
\small
\setlength{\tabcolsep}{3pt}
\begin{tabular}{@{}lccc@{}}
\toprule
Representation & PSNR$\uparrow$ & SSIM$\uparrow$ & SAM$\downarrow$ \\
\midrule
(i) Shared space, 28 per-band variables & 33.38 & 0.907 & 5.91 \\
(ii) 3DGS ($K=1$) & 34.96 & 0.929 & 5.99 \\
(iii) Ours ($K=3$) & 35.19 & 0.932 & 5.74 \\
\bottomrule
\end{tabular}
\end{table}

\begin{table}[t]
\centering
\footnotesize
\caption{Sensitivity to the number of spectral Gaussians per primitive, $K$, on KAIST (mean over 10 scenes and three seeds). All other settings are those of the final configuration; the main paper uses $K=3$. Only $K=1$ causes a clear drop, and the results are nearly constant for $K\geq3$.}
\label{tab:supp_sens_k}
\begin{tabular}{@{}lcccc@{}}
\toprule
$K$ & Params / primitive & PSNR & SSIM & SAM \\
\midrule
1 & 8 & 35.66 & 0.943 & 6.11 \\
2 & 11 & 36.67 & 0.954 & 5.00 \\
3 (ours) & 14 & 36.81 & 0.955 & 4.74 \\
4 & 17 & 36.87 & 0.955 & 4.69 \\
5 & 20 & 36.86 & 0.955 & 4.72 \\
6 & 23 & 36.86 & 0.954 & 4.77 \\
8 & 29 & 36.82 & 0.954 & 4.77 \\
14 & 47 & 36.81 & 0.953 & 4.77 \\
28 & 89 & 36.78 & 0.953 & 4.75 \\
\bottomrule
\end{tabular}
\end{table}

\begin{table}[t]
\centering
\footnotesize
\setlength{\tabcolsep}{3pt}
\caption{Sensitivity to the number of primitives on KAIST (mean over 10 scenes and three seeds). The initial number and the number below which merging stops are changed with the same ratio (3:1), with the same merging schedule; Final is the number at the end (mean over 30 runs). The main paper uses L.}
\label{tab:supp_sens_size}
\begin{tabular}{@{}lcccccc@{}}
\toprule
Size & Initial & Merging stops & Final & PSNR & SSIM & SAM \\
\midrule
S & 15k & 5k & 4.9k & 35.15 & 0.934 & 6.35 \\
M & 30k & 10k & 9.7k & 36.44 & 0.950 & 5.28 \\
L (ours) & 60k & 20k & 19.4k & 36.81 & 0.955 & 4.74 \\
\bottomrule
\end{tabular}
\end{table}


\begin{table}[t]
\centering
\footnotesize
\setlength{\tabcolsep}{3pt}
\caption{Arrangement and coverage of the primitives on KAIST (mean over 30 runs: 10 scenes and three seeds). $\bar d_{\mathrm{nn}}$: mean distance from each primitive center to its nearest neighbor. $R$: Clark--Evans index $\bar d_{\mathrm{nn}}/(0.5/\sqrt{\rho})$, where $\rho$ is the number of primitives per pixel ($R=1$ for a random arrangement and $R<1$ for a clustered one). $\bar\sigma$: median rendered width including the low-pass filter. $C<1$: fraction of pixels whose coverage $C(\mathbf p)=\sum_i\tilde G_i(\mathbf p)$ is below 1. Adding the Hessian ratio increases $\bar d_{\mathrm{nn}}$ and $R$ and decreases $C<1$ in all 30 paired runs.}
\label{tab:supp_prim_stats}
\begin{tabular}{@{}lcccc@{}}
\toprule
Configuration & $\bar d_{\mathrm{nn}}$ (px) & $R$ & $\bar\sigma$ (px) & $C<1$ (\%) \\
\midrule
Full (ours) & 1.01 & 1.06 & 1.59 & 0.36 \\
w/o ripple suppression & 1.05 & 1.10 & 1.53 & 0.15 \\
w/o Hessian ratio & 0.73 & 0.76 & 1.38 & 14.22 \\
w/o Hessian ratio \& ripple supp. & 0.81 & 0.84 & 1.21 & 16.98 \\
\bottomrule
\end{tabular}
\end{table}

\begin{table}[t]
\centering
\footnotesize
\setlength{\tabcolsep}{4pt}
\caption{Computation time, GPU memory, and number of parameters on KAIST scene 01 (unsupervised methods, one run at a time on a single NVIDIA RTX A6000). GPU memory is the peak memory allocated by PyTorch, and the number of parameters counts the values updated by the optimizer. For ours, which reduces the primitives from 60{,}000 to about 19{,}000 by merging, both are shown as beginning $\rightarrow$ end; the memory at the end is the peak over the last 1{,}000 iterations. The timed range differs slightly with the public code: the whole process for CEINR, and including the GAP-TV initialization for LRSDN. LRSDN was timed with our port of the authors' code using the same number of iterations as their public code. GSLR is our reimplementation (no public code) and uses the same custom CUDA rasterizer as ours.}
\label{tab:supp_profile}
\begin{tabular}{@{}lccc@{}}
\toprule
Method & Time (min) & GPU memory (GB) & \#Params \\
\midrule
LRSDN~\cite{chen2024hyperspectral} & 37.1 & 0.88 & 239{,}699 \\
CEINR~\cite{mei2024progressive} & 190.5 & 1.32 & 142{,}512 \\
LCTC~\cite{chen2025lctcpnplctc} & 7.1 & 1.18 & 510{,}224 \\
GSLR~\cite{zeng2026gaussian} & 42.3 & 0.09 & 1{,}201{,}350 \\
Ours & 14.1 & 2.84 $\rightarrow$ 0.62 & 840{,}000 $\rightarrow$ 272{,}426 \\
\bottomrule
\end{tabular}
\end{table}



\begin{table*}[p]
\centering
\footnotesize
\renewcommand{\arraystretch}{1.05}
\caption{All hyperparameters of CASSI-GS. The same values are used for every KAIST scene. The changes for the real data are listed in \cref{tab:hparams_real}.}
\label{tab:hparams}
\begin{tabularx}{\textwidth}{@{}>{\raggedright\arraybackslash}X >{\raggedright\arraybackslash}p{4.6cm}@{}}
\toprule
Hyperparameter & Value \\
\midrule
\multicolumn{2}{@{}l}{\textbf{Optimization}} \\
Iterations & 40{,}000 \\
Warm-up: fit only the spatial parameters and amplitudes to the panchromatic coarse estimate (spectral parameters frozen, no regularization) & 400 iterations \\
Optimizer & Adam ($\beta_1{=}0.9$, $\beta_2{=}0.999$, $\epsilon{=}10^{-8}$) \\
Learning rates (constant): position / log-scale / rotation & 0.08 / 0.015 / 0.01 \\
Learning rates (constant): amplitude (logit) / spectral center / log spectral width & 0.03 / 0.05 / 0.01 \\
Gradient clipping (global norm) & 1.0 \\
Random seeds (affect only the initial spectral centers) & 42, 43, 44 \\
\midrule
\multicolumn{2}{@{}l}{\textbf{Representation}} \\
Initial number of primitives & 60{,}000 (0.92 per pixel) \\
Number of spectral Gaussians per primitive, $K$ & 3 (14 parameters per primitive) \\
Initial spatial standard deviation & 2.0\,px \\
Initial spectral width & 6.0 bands \\
Initial component amplitude & $0.5/K$ \\
Initial component centers & uniform random over the bands \\
Range of the spatial standard deviation & $[0.05, 12]$\,px \\
Range of the spectral width & $[1, 50]$ bands \\
Allowed margin outside the image / outside the band range & 2\,px / 0.5 band \\
Coarse-to-fine: lower bound of the spatial standard deviation, decreased exponentially over the first 20{,}000 iterations (half of the total) & 2.0\,px $\rightarrow$ 0.05\,px \\
\midrule
\multicolumn{2}{@{}l}{\textbf{Rendering}} \\
Spatial low-pass width, lower bound & 0.1\,px \\
Spectral low-pass width $s_\lambda$ (added to the spectral variance, $(\sigma^\lambda_{ik})^2 + s_\lambda^2$, with the amplitude scaled by $\sigma^\lambda_{ik}/\sqrt{(\sigma^\lambda_{ik})^2 + s_\lambda^2}$) & 0.1 band \\
Ripple suppression: tolerance $\varepsilon_{\mathrm{r}}$ ($\tau \approx 0.551$) / cell size $c$ for the local spacing / upper bound of $\rho_i$ & 0.01 / 4\,px / 2\,px \\
Ripple suppression: the low-pass width $\rho_i$ is computed from the positions and widths with their gradients stopped & -- \\
\midrule
\multicolumn{2}{@{}l}{\textbf{Regularization}} \\
Hessian ratio: weight $w_{\mathrm{hr}}$ & $5\times10^{-4}$ \\
Hessian ratio: $\varepsilon_1$ (flat/edge threshold) / $\varepsilon_2$ (intensity offset) & 0.005 / 0.1 \\
Low rank: weight & $3\times10^{-2}$ \\
Low rank: rank, increased linearly over the iterations & $2 \rightarrow 6$ \\
Neighborhood spectral-shape consistency: weight / neighborhood width $\sigma_{\mathrm{n}}$ (pairs beyond $3\sigma_{\mathrm{n}}$ ignored) / applied every & 0.01 / 3\,px / 10 iterations \\
\midrule
\multicolumn{2}{@{}l}{\textbf{Merging}} \\
Schedule: every 500 iterations from iteration 2{,}000 to the end & 500 / 2{,}000 \\
Distance condition: center distance $\le$ $1.5\,\min(\bar{\sigma}_i, \bar{\sigma}_a)$ ($\bar{\sigma}=\sqrt{\sigma_x\sigma_y}$; $a$: anchor) & 1.5 \\
Size condition: $\max(\bar{\sigma}_i, \bar{\sigma}_a) / \min(\bar{\sigma}_i, \bar{\sigma}_a) \le 1.5$ & 1.5 \\
Spectral condition: cosine similarity of $\phi_i$ and $\phi_a$ (sampled at the 28 bands) $\ge 0.99$ & 0.99 \\
Grid cell size (one anchor per cell; recomputed at each merging step) & $3\times$ median of $\bar{\sigma}_i$ \\
Fraction of primitives removed per merging step, upper bound (beyond it, pairs with higher spectral cosine similarity first) & 5\% \\
Number of primitives below which merging stops & 20{,}000 (0.305 per pixel) \\
First moment of Adam for each merged primitive & reset to 0 (second moment kept) \\
Factor on the standard deviation of a merged primitive ($\approx\sqrt{2}$, keeps the coverage) & 1.41 \\
\midrule
\multicolumn{2}{@{}l}{\textbf{Forward model}} \\
Dispersion shift per band & 2\,px \\
\bottomrule
\end{tabularx}
\end{table*}

\begin{table*}[t]
\centering
\footnotesize
\renewcommand{\arraystretch}{1.05}
\caption{Changes for the real data. All other hyperparameters are those of \cref{tab:hparams}.}
\label{tab:hparams_real}
\begin{tabularx}{\textwidth}{@{}>{\raggedright\arraybackslash}X >{\raggedright\arraybackslash}p{4.6cm}@{}}
\toprule
Hyperparameter & Value \\
\midrule
Initial number of primitives (the real images are $660\times660$ pixels) & 400{,}000 (0.92 per pixel, as for KAIST) \\
Number of primitives below which merging stops (same density as KAIST) & 133{,}000  \\
Hessian ratio weight $w_{\mathrm{hr}}$ & $5\times10^{-3}$ \\
Mask: calibrated by the affine transform $\theta$ of SGDE~\cite{he2026sgde} & -- \\
Data term weighted by the per-pixel noise variance (shot noise and read noise; $\sigma_r = 5$\,DN, $\eta = 4$\,e$^-$/DN; full scale 4{,}091\,DN, 3{,}208\,DN for scene~5) & -- \\
Constant offset $c$ in $\hat{\meas}=\fwd_\theta(\cube)+c$: initial value = 1\% quantile of $\mathbf{Y}$, learned from iteration 10{,}000 with learning rate $10^{-4}$ & -- \\
Measurement scaling: 99.9th percentile of the coarse estimate set to 0.424 (median over the simulated KAIST measurements) & 0.424 \\
\bottomrule
\end{tabularx}
\end{table*}

\section{Analysis of CASSI-GS}
\label{sec:supp_analysis}

To examine the effects of spatial sharing and of the spectral representation, we compared the following three representations:
(i) a representation that shares the spatial shape across the 28 bands and holds the spectrum as 28 independent per-band values (a 28-channel version of RGB colors);
(ii) a 3D GS representation in which each primitive has a single Gaussian along the wavelength (without rotations that couple space and wavelength);
and (iii) a representation that shares the spatial shape and holds the spectrum as a sum of three Gaussians along the wavelength (ours).
To compare the representations themselves, we removed the neighborhood spectral-shape consistency and the merging from all three and used 60,000 primitives for each.
The other conditions (hyperparameters, scenes, and random seeds) were the same as in the ablation study.
The results are shown in \cref{tab:repr}.
The difference between (i) and (iii) reflects the effect of continuous spectra, and the difference between (ii) and (iii) reflects the effect of sharing one spatial shape among multiple spectral components. Our representation (iii) achieves the best PSNR, SSIM, and SAM among the three.

Table~\ref{tab:supp_sens_k} shows the sensitivity to the number of spectral Gaussians per primitive, $K$. With $K=1$, the accuracy drops clearly, and $K=2$ also lowers it slightly. For $K\geq3$, increasing $K$ hardly changes the accuracy. We therefore use $K=3$, the smallest value at which the accuracy saturates, which keeps the number of parameters small (14 per primitive).

Table~\ref{tab:supp_sens_size} shows the sensitivity to the number of primitives, where the initial number and the number below which merging stops are changed together at a ratio of 3:1 (S, M, and L). The accuracy increases with the number of primitives: it drops markedly for S, and M is still somewhat lower than L, which we use.

Figure~\ref{fig:supp_spectra} shows the spectra at ten pixels in KAIST scene 9. Although the peak positions differ among the chalk sticks, our method reproduces the spectra well at all ten pixels.

Figure~\ref{fig:supp_blotch} shows the effect of the regularizers on blotchy artifacts. Without both the Hessian ratio and ripple suppression, blotchy artifacts appear throughout the image. Adding ripple suppression (w/o Hessian ratio) reduces them to some extent, but not sufficiently. Replacing the Hessian ratio with TV~\cite{rudin1992nonlinear}, which is widely used to suppress spatial nonuniformity, improves the result considerably, but blotchy artifacts remain on the green chalk sticks. Our method suppresses the artifacts most effectively.

Table~\ref{tab:supp_prim_stats} shows the arrangement and coverage of the primitives on KAIST (10 scenes $\times$ three seeds, 30 runs in total). Adding the Hessian ratio (w/o Hessian ratio vs.\ Full) increases $\bar d_{\mathrm{nn}}$ from 0.73 to 1.01 pixels and changes the Clark--Evans index $R$ from 0.76 (clustered) to 1.06 (slightly more regular than random). Since the number of primitives is nearly the same in all configurations, the larger $\bar d_{\mathrm{nn}}$ indicates that the primitives are spread evenly instead of clustering. In addition, $\bar\sigma$ increases and $C<1$ decreases from 14.22\% to 0.36\%, so that coverage holes almost disappear. In contrast, adding ripple suppression without the Hessian ratio (w/o Hessian ratio \& ripple suppression vs.\ w/o Hessian ratio) widens $\bar\sigma$ through the low-pass filter and reduces $C<1$, but decreases $\bar d_{\mathrm{nn}}$ and $R$, so the arrangement does not become more even.

Table~\ref{tab:supp_profile} compares the computation time, GPU memory, and number of parameters of the unsupervised methods. The parameters are the network weights for LRSDN, CEINR, and LCTC, and the parameters of the primitives (and the basis) for GSLR and ours. GSLR has the most parameters because it keeps 60,000 2D Gaussians without merging, each with coefficients for $R=15$ bases, and it uses the least GPU memory. Our method uses the most GPU memory at the beginning (2.84\,GB with 60,000 primitives), but as merging reduces the primitives to about 19,000, the memory decreases to 0.62\,GB and the number of parameters from 840,000 to 272,426. LCTC is the fastest, followed by ours.

\section*{Additional Reference}
{\small
\begin{list}{}{%
  \settowidth{\labelwidth}{[S1]}%
  \setlength{\leftmargin}{\dimexpr\labelwidth+\labelsep\relax}%
  \setlength{\itemindent}{0pt}\setlength{\topsep}{0pt}}
\item[{[S1]}] Alessandro Foi, Mejdi Trimeche, Vladimir Katkovnik, and Karen Egiazarian.
Practical Poissonian-Gaussian Noise Modeling and Fitting for Single-Image Raw-Data.
\emph{IEEE Transactions on Image Processing}, 17(10):1737--1754, 2008.
\end{list}}

%% file: refs.bib
@article{bhargava2024hyperspectral,
  title = {Hyperspectral {{Imaging}} and {{Its Applications}}: {{A Review}}},
  author = {Bhargava, Anuja and Sachdeva, Ashish and Sharma, Kulbhushan and Alsharif, Mohammed H. and Uthansakul, Peerapong and Uthansakul, Monthippa},
  year = 2024,
  journal = {Heliyon},
  volume = {10},
  number = {12},
  pages = {e33208},
  doi = {10.1016/j.heliyon.2024.e33208}
}

@article{bioucas-dias2007new,
  title = {A {{New TwIST}}: {{Two-Step Iterative Shrinkage}}/{{Thresholding Algorithms}} for {{Image Restoration}}},
  shorttitle = {A {{New TwIST}}},
  author = {{Bioucas-Dias}, J.M. and Figueiredo, M.A.T.},
  year = 2007,
  month = dec,
  journal = {IEEE Transactions on Image Processing},
  volume = {16},
  number = {12},
  pages = {2992--3004},
  issn = {1057-7149, 1941-0042},
  doi = {10.1109/TIP.2007.909319},
  urldate = {2026-11-01},
  copyright = {https://ieeexplore.ieee.org/Xplorehelp/downloads/license-information/IEEE.html},
  langid = {english}
}

@article{bredies2010tgv,
  title = {Total {{Generalized Variation}}},
  author = {Bredies, Kristian and Kunisch, Karl and Pock, Thomas},
  year = 2010,
  journal = {SIAM Journal on Imaging Sciences},
  volume = {3},
  number = {3},
  pages = {492--526},
  doi = {10.1137/090769521}
}

@incollection{cai2022coarsetofine,
  title = {Coarse-to-{{Fine Sparse Transformer}} for {{Hyperspectral Image Reconstruction}}},
  booktitle = {Computer {{Vision}} -- {{ECCV}} 2022},
  author = {Cai, Yuanhao and Lin, Jing and Hu, Xiaowan and Wang, Haoqian and Yuan, Xin and Zhang, Yulun and Timofte, Radu and Van Gool, Luc},
  editor = {Avidan, Shai and Brostow, Gabriel and Ciss{\'e}, Moustapha and Farinella, Giovanni Maria and Hassner, Tal},
  year = 2022,
  series = {Lecture {{Notes}} in {{Computer Science}}},
  volume = {13677},
  pages = {686--704},
  publisher = {Springer Nature Switzerland},
  address = {Cham},
  doi = {10.1007/978-3-031-19790-1_41},
  urldate = {2026-11-01},
  isbn = {978-3-031-19789-5 978-3-031-19790-1},
  langid = {english}
}

@inproceedings{cai2022degradationaware,
  title = {Degradation-{{Aware Unfolding Half-Shuffle Transformer}} for {{Spectral Compressive Imaging}}},
  booktitle = {Advances in {{Neural Information Processing Systems}}},
  author = {Cai, Yuanhao and Lin, Jing and Wang, Haoqian and Yuan, Xin and Ding, Henghui and Zhang, Yulun and Timofte, Radu and Van Gool, Luc},
  year = 2022,
  volume = {35},
  pages = {37749--37761},
  publisher = {Curran Associates, Inc.},
  doi = {10.52202/068431-2736},
  urldate = {2026-11-01}
}

@inproceedings{cai2022maskguided,
  title = {Mask-Guided {{Spectral}}-Wise {{Transformer}} for {{Efficient Hyperspectral Image Reconstruction}}},
  booktitle = {2022 {{IEEE}}/{{CVF Conference}} on {{Computer Vision}} and {{Pattern Recognition}} ({{CVPR}})},
  author = {Cai, Yuanhao and Lin, Jing and Hu, Xiaowan and Wang, Haoqian and Yuan, Xin and Zhang, Yulun and Timofte, Radu and Van Gool, Luc},
  year = 2022,
  month = jun,
  pages = {17481--17490},
  publisher = {IEEE},
  address = {New Orleans, LA, USA},
  doi = {10.1109/CVPR52688.2022.01698},
  urldate = {2026-11-01},
  copyright = {https://doi.org/10.15223/policy-029},
  isbn = {978-1-6654-6946-3},
  langid = {english}
}

@inproceedings{chen2024hac,
  title = {{{HAC}}: {{Hash-Grid Assisted Context}} for {{3D Gaussian Splatting Compression}}},
  booktitle = {Computer {{Vision}} -- {{ECCV}} 2024},
  author = {Chen, Yihang and Wu, Qianyi and Lin, Weiyao and Harandi, Mehrtash and Cai, Jianfei},
  year = 2024,
  series = {Lecture {{Notes}} in {{Computer Science}}},
  volume = {15065},
  pages = {422--438},
  publisher = {Springer Nature Switzerland},
  address = {Cham},
  doi = {10.1007/978-3-031-72667-5_24}
}

@article{chen2024hyperspectral,
  title = {Hyperspectral {{Compressive Snapshot Reconstruction}} via {{Coupled Low-Rank Subspace Representation}} and {{Self-Supervised Deep Network}}},
  author = {Chen, Yong and Lai, Wenzhen and He, Wei and Zhao, Xi-Le and Zeng, Jinshan},
  year = 2024,
  journal = {IEEE Transactions on Image Processing},
  volume = {33},
  pages = {926--941},
  issn = {1057-7149, 1941-0042},
  doi = {10.1109/TIP.2024.3354127},
  urldate = {2026-11-01},
  copyright = {https://ieeexplore.ieee.org/Xplorehelp/downloads/license-information/IEEE.html},
  langid = {english}
}

@article{chen2024spectralwise,
  title = {Spectral-{{Wise Implicit Neural Representation}} for {{Hyperspectral Image Reconstruction}}},
  author = {Chen, Huan and Zhao, Wangcai and Xu, Tingfa and Shi, Guokai and Zhou, Shiyun and Liu, Peifu and Li, Jianan},
  year = 2024,
  month = may,
  journal = {IEEE Transactions on Circuits and Systems for Video Technology},
  volume = {34},
  number = {5},
  pages = {3714--3727},
  issn = {1051-8215, 1558-2205},
  doi = {10.1109/TCSVT.2023.3318366},
  urldate = {2026-11-01},
  copyright = {https://ieeexplore.ieee.org/Xplorehelp/downloads/license-information/IEEE.html}
}

@inproceedings{chen2025generalized,
  title = {Generalized and {{Efficient 2D Gaussian Splatting}} for {{Arbitrary-Scale Super-Resolution}}},
  booktitle = {2025 {{IEEE}}/{{CVF International Conference}} on {{Computer Vision}} ({{ICCV}})},
  author = {Chen, Du and Chen, Liyi and Zhang, Zhengqiang and Zhang, Lei},
  year = 2025,
  month = oct,
  pages = {26435--26445},
  publisher = {IEEE},
  address = {Honolulu, HI, USA},
  doi = {10.1109/ICCV51701.2025.02453},
  urldate = {2026-11-01},
  copyright = {https://doi.org/10.15223/policy-029},
  isbn = {979-8-3315-8775-8}
}

@article{chen2025lctcpnplctc,
  title = {{{LCTC}}: {{Lightweight Convolutional Thresholding Sparse Coding Network Prior}} for {{Compressive Hyperspectral Imaging}}},
  author = {Chen, Yurong and Wang, Yaonan and Wang, Xiaodong and Yuan, Xin and Zhang, Hui},
  year = 2025,
  journal = {IEEE Transactions on Image Processing},
  volume = {34},
  pages = {4286--4301},
  doi = {10.1109/TIP.2025.3583951}
}

@article{chen2025ldmeet,
  title = {Low-{{Rank Tensor Meets Deep Prior}}: {{Coupling Model-Driven}} and {{Data-Driven Methods}} for {{Hyperspectral Image Reconstruction}}},
  author = {Chen, Yong and Yuan, Feiwang and Lai, Wenzhen and Zeng, Jinshan and He, Wei and Huang, Qing},
  year = 2025,
  journal = {IEEE Transactions on Circuits and Systems for Video Technology},
  volume = {35},
  number = {11},
  pages = {11685--11697},
  doi = {10.1109/TCSVT.2025.3575470}
}

@inproceedings{chen2026localityaware,
  title = {Locality-{{Aware Density Control}} for {{Efficient Gaussian}}-Based {{Image Representation}}},
  booktitle = {Proceedings of the 34th {{ACM International Conference}} on {{Multimedia}}},
  author = {Chen, Jiacong and Mao, Qingyu and Meng, Xiandong and Liu, Shuai and Li, Chao and Meng, Fanyang and Bao, Youneng and Liang, Yongsheng},
  year = 2026,
  eprint = {2607.17896},
  pages = {438--446},
  publisher = {ACM},
  address = {New York, NY, USA},
  doi = {10.1145/3767308.3835325},
  urldate = {2026-11-01},
  archiveprefix = {arXiv},
  langid = {english}
}

@inproceedings{chen2026phycosf,
  title = {Phy-{{CoSF}}: {{Physics-Guided Continuous Spectral Fields Reconstruction}} and {{Spectral Super-Resolution}} for {{Snapshot Compressive Imaging}}},
  shorttitle = {Phy-{{CoSF}}},
  booktitle = {Proceedings of the 43rd {{International Conference}} on {{Machine Learning}}},
  author = {Chen, Wudi and Zha, Zhiyuan and Yuan, Xin and Wang, Shigang and Wen, Bihan and Zhou, Jiantao and Yan, Gang and Fan, Zipei and Zhu, Ce},
  year = 2026,
  series = {Proceedings of {{Machine Learning Research}}},
  volume = {306},
  eprint = {2605.13583},
  primaryclass = {cs.CV},
  pages = {18041--18055},
  publisher = {PMLR},
  urldate = {2026-11-01},
  archiveprefix = {arXiv},
  langid = {english}
}

@article{choi2017highquality,
  title = {{{High-Quality Hyperspectral Reconstruction Using a Spectral Prior}}},
  author = {Choi, Inchang and Jeon, Daniel S. and Nam, Giljoo and Gutierrez, Diego and Kim, Min H.},
  year = 2017,
  month = dec,
  journal = {ACM Transactions on Graphics},
  volume = {36},
  number = {6},
  pages = {218:1--218:13},
  issn = {0730-0301, 1557-7368},
  doi = {10.1145/3130800.3130810},
  urldate = {2026-11-01},
  langid = {english}
}

@article{clark1954distance,
  title = {Distance to {{Nearest Neighbor}} as a {{Measure}} of {{Spatial Relationships}} in {{Populations}}},
  author = {Clark, Philip J. and Evans, Francis C.},
  year = 1954,
  month = oct,
  journal = {Ecology},
  volume = {35},
  number = {4},
  pages = {445--453},
  issn = {00129658},
  doi = {10.2307/1931034},
  urldate = {2026-11-13},
  copyright = {http://doi.wiley.com/10.1002/tdm\_license\_1.1},
  langid = {english}
}

@inproceedings{dong2023residual,
  title = {Residual {{Degradation Learning Unfolding Framework}} with {{Mixing Priors Across Spectral}} and {{Spatial}} for {{Compressive Spectral Imaging}}},
  booktitle = {2023 {{IEEE}}/{{CVF Conference}} on {{Computer Vision}} and {{Pattern Recognition}} ({{CVPR}})},
  author = {Dong, Yubo and Gao, Dahua and Qiu, Tian and Li, Yuyan and Yang, Minxi and Shi, Guangming},
  year = 2023,
  month = jun,
  pages = {22262--22271},
  publisher = {IEEE},
  address = {Vancouver, BC, Canada},
  doi = {10.1109/CVPR52729.2023.02132},
  urldate = {2026-11-01},
  copyright = {https://doi.org/10.15223/policy-029},
  isbn = {979-8-3503-0129-8}
}

@misc{fang20264dsg,
  title = {Shared-{{Structure 4D Spectral Gaussian Representation}} for {{Sparse-View Spectral CT Reconstruction}}},
  author = {Fang, Jiancheng and Wang, Shaoyu and Xia, Wenjun and Chen, Yang and Liu, Qiegen},
  year = 2026,
  number = {arXiv:2608.16463},
  eprint = {2608.16463},
  publisher = {arXiv},
  doi = {10.48550/arXiv.2608.16463},
  archiveprefix = {arXiv},
  howpublished = {arXiv preprint arXiv:2608.16463}
}

@inproceedings{frangi1998multiscale,
  title = {Multiscale {{Vessel Enhancement Filtering}}},
  booktitle = {Medical {{Image Computing}} and {{Computer-Assisted Intervention}} --- {{MICCAI}}'98},
  author = {Frangi, Alejandro F. and Niessen, Wiro J. and Vincken, Koen L. and Viergever, Max A.},
  year = 1998,
  series = {Lecture {{Notes}} in {{Computer Science}}},
  volume = {1496},
  pages = {130--137},
  publisher = {Springer Berlin Heidelberg},
  address = {Berlin, Heidelberg},
  doi = {10.1007/BFb0056195}
}

@inproceedings{he2026sgde,
  title = {{{SGDE}}: {{Self}}-Supervised {{Geometry Degradation Estimation Framework}} for {{Coded Aperture Compressive Spectral Imaging}}},
  booktitle = {Proceedings of the {{IEEE}}/{{CVF Conference}} on {{Computer Vision}} and {{Pattern Recognition}} ({{CVPR}})},
  author = {He, Yuqiao and Liu, Xiaoyan and Mao, Jianxu and Wang, Yaonan and Zhang, Hui and Liu, Lizhu and Chen, Yurong and He, Wenbin},
  year = 2026,
  pages = {34084--34094},
  langid = {english}
}

@article{hong2026hyperspectral,
  title = {Hyperspectral {{Imaging}}},
  author = {Hong, Danfeng and Li, Chenyu and Yokoya, Naoto and Zhang, Bing and Jia, Xiuping and Plaza, Antonio and Gamba, Paolo and Benediktsson, Jon Atli and Chanussot, Jocelyn},
  year = 2026,
  journal = {Nature Reviews Methods Primers},
  volume = {6},
  number = {1},
  pages = {19},
  doi = {10.1038/s43586-026-00470-x}
}

@article{hu2025gaussiansr,
  title = {{{GaussianSR}}: {{High Fidelity 2D Gaussian Splatting}} for {{Arbitrary-Scale Image Super-Resolution}}},
  shorttitle = {{{GaussianSR}}},
  author = {Hu, Jintong and Xia, Bin and Chen, Bin and Yang, Wenming and Zhang, Lei},
  year = 2025,
  month = apr,
  journal = {Proceedings of the AAAI Conference on Artificial Intelligence},
  volume = {39},
  number = {4},
  pages = {3554--3562},
  issn = {2374-3468, 2159-5399},
  doi = {10.1609/aaai.v39i4.32369},
  urldate = {2026-11-01}
}

@inproceedings{huang2026lrdun,
  title = {{{LRDUN}}: {{A Low-Rank Deep Unfolding Network}} for {{Efficient Spectral Compressive Imaging}}},
  shorttitle = {{{LRDUN}}},
  booktitle = {Proceedings of the {{IEEE}}/{{CVF Conference}} on {{Computer Vision}} and {{Pattern Recognition}}},
  author = {Huang, He and Guo, Yujun and He, Wei},
  year = 2026,
  pages = {10556--10566},
  urldate = {2026-11-01},
  langid = {english}
}

@inproceedings{hyung2024effectiverank,
  title = {Effective {{Rank Analysis}} and {{Regularization}} for {{Enhanced 3D Gaussian Splatting}}},
  booktitle = {Advances in {{Neural Information Processing Systems}} ({{NeurIPS}})},
  author = {Hyung, Junha and Hong, Susung and Hwang, Sungwon and Lee, Jaeseong and Choo, Jaegul and Kim, Jin-Hwa},
  year = 2024,
  volume = {37},
  pages = {110412--110435},
  doi = {10.52202/079017-3505}
}

@article{jin2025compressive,
  title = {Compressive {{Imaging Reconstruction}} via {{Tensor Decomposition-Based Multi-Resolution Grid Encoding}}},
  author = {Jin, Zhenyu and Luo, Yisi and Zhao, Xile and Meng, Deyu},
  year = 2026,
  journal = {Inverse Problems and Imaging},
  eprint = {2507.07707},
  primaryclass = {eess.IV},
  publisher = {arXiv},
  doi = {10.3934/ipi.2026076},
  urldate = {2026-11-01},
  archiveprefix = {arXiv},
  langid = {english}
}

@article{kerbl20233d,
  title = {{{3D Gaussian Splatting}} for {{Real-Time Radiance Field Rendering}}},
  author = {Kerbl, Bernhard and Kopanas, Georgios and Leimkuehler, Thomas and Drettakis, George},
  year = 2023,
  month = aug,
  journal = {ACM Transactions on Graphics},
  volume = {42},
  number = {4},
  pages = {1--14},
  issn = {0730-0301, 1557-7368},
  doi = {10.1145/3592433},
  urldate = {2026-11-01},
  langid = {english}
}

@article{kerbl2024hierarchical,
  title = {A {{Hierarchical 3D Gaussian Representation}} for {{Real-Time Rendering}} of {{Very Large Datasets}}},
  author = {Kerbl, Bernhard and Meuleman, Andreas and Kopanas, Georgios and Wimmer, Michael and Lanvin, Alexandre and Drettakis, George},
  year = 2024,
  journal = {ACM Transactions on Graphics},
  volume = {43},
  number = {4},
  pages = {62:1--62:15},
  doi = {10.1145/3658160}
}

@inproceedings{kingma2015adam,
  title = {Adam: {{A Method}} for {{Stochastic Optimization}}},
  booktitle = {International {{Conference}} on {{Learning Representations}} ({{ICLR}})},
  author = {Kingma, Diederik P. and Ba, Jimmy},
  year = 2015
}

@techreport{kokaly2017usgs,
  type = {Data {{Series}}},
  title = {{{USGS Spectral Library Version 7}}},
  author = {Kokaly, Raymond F. and Clark, Roger N. and Swayze, Gregg A. and Livo, K. Eric and Hoefen, Todd M. and Pearson, Neil C. and Wise, Richard A. and Benzel, William M. and Lowers, Heather A. and Driscoll, Rhonda L. and Klein, Anna J.},
  year = 2017,
  number = {1035},
  address = {Reston, VA},
  institution = {U.S. Geological Survey},
  doi = {10.3133/ds1035}
}

@article{kruse1993spectral,
  title = {{{The Spectral Image Processing System (SIPS)---Interactive Visualization and Analysis of Imaging Spectrometer Data}}},
  author = {Kruse, F.A. and Lefkoff, A.B. and Boardman, J.W. and Heidebrecht, K.B. and Shapiro, A.T. and Barloon, P.J. and Goetz, A.F.H.},
  year = 1993,
  month = may,
  journal = {Remote Sensing of Environment},
  volume = {44},
  number = {2-3},
  pages = {145--163},
  issn = {00344257},
  doi = {10.1016/0034-4257(93)90013-N},
  urldate = {2026-11-11},
  copyright = {https://www.elsevier.com/tdm/userlicense/1.0/},
  langid = {english}
}

@inproceedings{lee2024compact,
  title = {Compact {{3D Gaussian Representation}} for {{Radiance Field}}},
  booktitle = {2024 {{IEEE}}/{{CVF Conference}} on {{Computer Vision}} and {{Pattern Recognition}} ({{CVPR}})},
  author = {Lee, Joo Chan and Rho, Daniel and Sun, Xiangyu and Ko, Jong Hwan and Park, Eunbyung},
  year = 2024,
  month = jun,
  pages = {21719--21728},
  publisher = {IEEE},
  address = {Seattle, WA, USA},
  doi = {10.1109/CVPR52733.2024.02052},
  urldate = {2026-11-01},
  copyright = {https://doi.org/10.15223/policy-029},
  isbn = {979-8-3503-5300-6}
}

@article{lefkimmiatis2013hessian,
  title = {Hessian {{Schatten-Norm Regularization}} for {{Linear Inverse Problems}}},
  author = {Lefkimmiatis, Stamatios and Ward, John Paul and Unser, Michael},
  year = 2013,
  month = may,
  journal = {IEEE Transactions on Image Processing},
  volume = {22},
  number = {5},
  pages = {1873--1888},
  issn = {1057-7149, 1941-0042},
  doi = {10.1109/TIP.2013.2237919},
  urldate = {2026-11-01},
  copyright = {https://ieeexplore.ieee.org/Xplorehelp/downloads/license-information/IEEE.html}
}

@article{li2025lorsgs,
  title = {{{LoR-SGS}}: {{Hyperspectral Image Compression}} via {{Low-Rank Spectral Gaussian Splatting}}},
  author = {Li, Tianyu and Wang, Ting and Zhao, Xile and Wang, Chao},
  year = 2025,
  journal = {IEEE Transactions on Geoscience and Remote Sensing},
  volume = {63},
  pages = {4707913},
  doi = {10.1109/TGRS.2025.3623253}
}

@article{li2026gaussianimage,
  title = {{{GaussianImage}}++: {{Boosted Image Representation}} and {{Compression}} with {{2D Gaussian Splatting}}},
  shorttitle = {{{GaussianImage}}++},
  author = {Li, Tiantian and Zhang, Xinjie and Ge, Xingtong and Xu, Tongda and He, Dailan and Zhang, Jun and Wang, Yan},
  year = 2026,
  month = mar,
  journal = {Proceedings of the AAAI Conference on Artificial Intelligence},
  volume = {40},
  number = {8},
  pages = {6442--6449},
  issn = {2374-3468, 2159-5399},
  doi = {10.1609/aaai.v40i8.37572},
  urldate = {2026-11-01}
}

@article{liu2019rank,
  title = {Rank {{Minimization}} for {{Snapshot Compressive Imaging}}},
  author = {Liu, Yang and Yuan, Xin and Suo, Jinli and Brady, David J. and Dai, Qionghai},
  year = 2019,
  month = dec,
  journal = {IEEE Transactions on Pattern Analysis and Machine Intelligence},
  volume = {41},
  number = {12},
  pages = {2990--3006},
  issn = {0162-8828, 2160-9292, 1939-3539},
  doi = {10.1109/TPAMI.2018.2873587},
  urldate = {2026-11-01},
  copyright = {https://ieeexplore.ieee.org/Xplorehelp/downloads/license-information/IEEE.html}
}

@article{liu2025gshsi,
  title = {{{GS-HSI}}: Cross-Band Prior-Guided {{2D Gaussian}} Splatting for Hyperspectral Image Compression},
  shorttitle = {{{GS-HSI}}},
  author = {Liu, Jiahui and Zhang, Lili and Gao, Yang and Qu, Lele},
  year = 2025,
  journal = {Chinese Optics Letters},
  volume = {23},
  number = {11},
  pages = {111106},
  issn = {1671-7694},
  doi = {10.3788/COL202523.111106},
  urldate = {2026-11-01},
  langid = {english}
}

@article{mei2024progressive,
  title = {Progressive {{Self-Supervised Learning}} for {{CASSI Computational Spectral Cameras}}},
  author = {Mei, Xiaoyin and Li, Yuqi and Fu, Qiang and Heidrich, Wolfgang},
  year = 2024,
  journal = {IEEE Transactions on Computational Imaging},
  volume = {10},
  pages = {1505--1518},
  issn = {2333-9403, 2334-0118, 2573-0436},
  doi = {10.1109/TCI.2024.3463478},
  urldate = {2026-11-01},
  copyright = {https://ieeexplore.ieee.org/Xplorehelp/downloads/license-information/IEEE.html},
  langid = {english}
}

@incollection{meng2020endtoend,
  title = {End-to-{{End Low Cost Compressive Spectral Imaging}} with {{Spatial-Spectral Self-Attention}}},
  booktitle = {Computer {{Vision}} -- {{ECCV}} 2020},
  author = {Meng, Ziyi and Ma, Jiawei and Yuan, Xin},
  editor = {Vedaldi, Andrea and Bischof, Horst and Brox, Thomas and Frahm, Jan-Michael},
  year = 2020,
  series = {Lecture {{Notes}} in {{Computer Science}}},
  volume = {12368},
  pages = {187--204},
  publisher = {Springer International Publishing},
  address = {Cham},
  doi = {10.1007/978-3-030-58592-1_12},
  urldate = {2026-11-01},
  isbn = {978-3-030-58591-4 978-3-030-58592-1},
  langid = {english}
}

@inproceedings{meng2021selfsupervised,
  title = {{{Self-Supervised Neural Networks for Spectral Snapshot Compressive Imaging}}},
  booktitle = {2021 {{IEEE}}/{{CVF International Conference}} on {{Computer Vision}} ({{ICCV}})},
  author = {Meng, Ziyi and Yu, Zhenming and Xu, Kun and Yuan, Xin},
  year = 2021,
  month = oct,
  pages = {2602--2611},
  publisher = {IEEE},
  address = {Montreal, QC, Canada},
  doi = {10.1109/ICCV48922.2021.00262},
  urldate = {2026-11-01},
  copyright = {https://doi.org/10.15223/policy-029},
  isbn = {978-1-6654-2812-5},
  langid = {english}
}

@inproceedings{morgenstern2024sog,
  title = {Compact {{3D Scene Representation}} via {{Self-Organizing Gaussian Grids}}},
  booktitle = {Computer {{Vision}} -- {{ECCV}} 2024},
  author = {Morgenstern, Wieland and Barthel, Florian and Hilsmann, Anna and Eisert, Peter},
  year = 2024,
  series = {Lecture {{Notes}} in {{Computer Science}}},
  volume = {15143},
  pages = {18--34},
  publisher = {Springer Nature Switzerland},
  address = {Cham},
  doi = {10.1007/978-3-031-73013-9_2}
}

@inproceedings{narayanan2026diffusiondenoised,
  title = {Diffusion-{{Denoised Hyperspectral Gaussian Splatting}}},
  booktitle = {2026 {{International Conference}} on {{3D Vision}} ({{3DV}})},
  author = {Narayanan, Sunil Kumar and Zhao, Lingjun and Gan, Lu and Chen, Yongsheng},
  year = 2026,
  month = mar,
  pages = {1218--1227},
  publisher = {IEEE},
  address = {Vancouver, BC, Canada},
  doi = {10.1109/3DV69130.2026.00120},
  urldate = {2026-11-01},
  copyright = {https://doi.org/10.15223/policy-029},
  isbn = {979-8-3315-7312-6}
}

@inproceedings{park2025dropgaussian,
  title = {{{DropGaussian}}: {{Structural Regularization}} for {{Sparse}}-View {{Gaussian Splatting}}},
  booktitle = {Proceedings of the {{IEEE}}/{{CVF Conference}} on {{Computer Vision}} and {{Pattern Recognition}} ({{CVPR}})},
  author = {Park, Hyunwoo and Ryu, Gun and Kim, Wonjun},
  year = 2025,
  pages = {21600--21609},
  doi = {10.1109/CVPR52734.2025.02012}
}

@inproceedings{paszke2019pytorcha,
  title = {{{PyTorch}}: {{An Imperative Style}}, {{High-Performance Deep Learning Library}}},
  shorttitle = {{{PyTorch}}},
  booktitle = {Advances in {{Neural Information Processing Systems}}},
  author = {Paszke, Adam and Gross, Sam and Massa, Francisco and Lerer, Adam and Bradbury, James and Chanan, Gregory and Killeen, Trevor and Lin, Zeming and Gimelshein, Natalia and Antiga, Luca and Desmaison, Alban and Kopf, Andreas and Yang, Edward and DeVito, Zachary and Raison, Martin and Tejani, Alykhan and Chilamkurthy, Sasank and Steiner, Benoit and Fang, Lu and Bai, Junjie and Chintala, Soumith},
  year = 2019,
  volume = {32},
  publisher = {Curran Associates, Inc.},
  urldate = {2026-11-11}
}

@inproceedings{peng2026pixel,
  title = {Pixel to {{Gaussian}}: {{Ultra-Fast Continuous Super-Resolution}} with {{2D Gaussian Modeling}}},
  booktitle = {International {{Conference}} on {{Learning Representations}} ({{ICLR}})},
  author = {Peng, Long and Wu, Anran and Li, Wenbo and Xia, Peizhe and Zhang, Xinjie and Dai, Xueyuan and Di, Xin and Sun, Haoze and Pei, Renjing and Wang, Yang and Cao, Yang and Zha, Zheng-Jun},
  year = 2026,
  urldate = {2026-11-01},
  langid = {english}
}

@article{rudin1992nonlinear,
  title = {Nonlinear {{Total Variation Based Noise Removal Algorithms}}},
  author = {Rudin, Leonid I. and Osher, Stanley and Fatemi, Emad},
  year = 1992,
  journal = {Physica D: Nonlinear Phenomena},
  volume = {60},
  number = {1--4},
  pages = {259--268},
  doi = {10.1016/0167-2789(92)90242-F}
}

@article{runnalls2007kullback,
  title = {Kullback-{{Leibler Approach}} to {{Gaussian Mixture Reduction}}},
  author = {Runnalls, Andrew R.},
  year = 2007,
  journal = {IEEE Transactions on Aerospace and Electronic Systems},
  volume = {43},
  number = {3},
  pages = {989--999},
  doi = {10.1109/TAES.2007.4383588}
}

@inproceedings{song2026d2gs,
  title = {{{D}}{$^{2}$}{{GS}}: {{Depth}}-and-{{Density Guided Gaussian Splatting}} for {{Stable}} and {{Accurate Sparse-View Reconstruction}}},
  booktitle = {International {{Conference}} on {{Learning Representations}} ({{ICLR}})},
  author = {Song, Meixi and Lin, Xin and Zhang, Dizhe and Li, Haodong and Li, Xiangtai and Du, Bo and Qi, Lu},
  year = 2026
}

@article{teng2026dbp,
  title = {Deep {{Balanced Prior}} for {{Refined Self-Supervised Reconstruction}} of {{Snapshot Spectral Compressive Imaging}}},
  author = {Teng, Feifan and Jia, Yizhen and An, Ping and Shen, Liquan and Wang, Shuming and Yuan, Xin and Hua, Xia},
  year = 2026,
  journal = {IEEE Transactions on Instrumentation and Measurement},
  volume = {75},
  pages = {5009914},
  doi = {10.1109/TIM.2026.3693425}
}

@inproceedings{thirgood2025hypergs,
  title = {{{HyperGS}}: {{Hyperspectral 3D Gaussian Splatting}}},
  shorttitle = {{{HyperGS}}},
  booktitle = {2025 {{IEEE}}/{{CVF Conference}} on {{Computer Vision}} and {{Pattern Recognition}} ({{CVPR}})},
  author = {Thirgood, Christopher and Mendez, Oscar and Ling, Erin and Storey, Jon and Hadfield, Simon},
  year = 2025,
  month = jun,
  pages = {5970--5979},
  publisher = {IEEE},
  address = {Nashville, TN, USA},
  doi = {10.1109/CVPR52734.2025.00560},
  urldate = {2026-11-01},
  copyright = {https://doi.org/10.15223/policy-029},
  isbn = {979-8-3315-4364-8}
}

@article{tsai1998derivative,
  title = {Derivative {{Analysis}} of {{Hyperspectral Data}}},
  author = {Tsai, Fuan and Philpot, William},
  year = 1998,
  month = oct,
  journal = {Remote Sensing of Environment},
  volume = {66},
  number = {1},
  pages = {41--51},
  issn = {00344257},
  doi = {10.1016/S0034-4257(98)00032-7},
  urldate = {2026-11-11},
  langid = {english}
}

@article{ulyanov2020deep,
  title = {Deep {{Image Prior}}},
  author = {Ulyanov, Dmitry and Vedaldi, Andrea and Lempitsky, Victor},
  year = 2020,
  month = jul,
  journal = {International Journal of Computer Vision},
  volume = {128},
  number = {7},
  pages = {1867--1888},
  issn = {0920-5691, 1573-1405},
  doi = {10.1007/s11263-020-01303-4},
  urldate = {2026-11-01},
  langid = {english}
}

@article{wagadarikar2008single,
  title = {Single Disperser Design for Coded Aperture Snapshot Spectral Imaging},
  author = {Wagadarikar, Ashwin and John, Renu and Willett, Rebecca and Brady, David},
  year = 2008,
  month = apr,
  journal = {Applied Optics},
  volume = {47},
  number = {10},
  pages = {B44--B51},
  issn = {0003-6935, 1539-4522},
  doi = {10.1364/AO.47.000B44},
  urldate = {2026-11-01},
  copyright = {https://doi.org/10.1364/OA\_License\_v1\#VOR},
  langid = {english}
}

@incollection{wang2022modeling,
  title = {Modeling {{Mask Uncertainty}} in {{Hyperspectral Image Reconstruction}}},
  booktitle = {Computer {{Vision}} -- {{ECCV}} 2022},
  author = {Wang, Jiamian and Zhang, Yulun and Yuan, Xin and Meng, Ziyi and Tao, Zhiqiang},
  editor = {Avidan, Shai and Brostow, Gabriel and Ciss{\'e}, Moustapha and Farinella, Giovanni Maria and Hassner, Tal},
  year = 2022,
  series = {Lecture {{Notes}} in {{Computer Science}}},
  volume = {13679},
  pages = {112--129},
  publisher = {Springer Nature Switzerland},
  address = {Cham},
  doi = {10.1007/978-3-031-19800-7_7},
  urldate = {2026-11-01},
  isbn = {978-3-031-19799-4 978-3-031-19800-7},
  langid = {english}
}

@inproceedings{wang2025scigs,
  title = {{{SCIGS}}: {{3D Gaussians Splatting}} from {{A Snapshot Compressive Image}}},
  shorttitle = {{{SCIGS}}},
  booktitle = {2025 {{IEEE International Conference}} on {{Image Processing}} ({{ICIP}})},
  author = {Wang, Zixu and Yang, Hao and Guo, Yu and Wang, Fei},
  year = 2025,
  month = sep,
  pages = {1013--1018},
  publisher = {IEEE},
  address = {Anchorage, AK, USA},
  doi = {10.1109/ICIP55913.2025.11084292},
  urldate = {2026-11-01},
  copyright = {https://doi.org/10.15223/policy-029},
  isbn = {979-8-3315-2379-4}
}

@inproceedings{wu2025discretized,
  title = {Discretized {{Gaussian Representation}} for {{Tomographic Reconstruction}}},
  booktitle = {2025 {{IEEE}}/{{CVF International Conference}} on {{Computer Vision}} ({{ICCV}})},
  author = {Wu, Shaokai and Lu, Yuxiang and Guo, Yapan and Ji, Wei and Huang, Suizhi and Yang, Fengyu and Sirejiding, Shalayiding and He, Qichen and Tong, Jing and Ji, Yanbiao and Ding, Yue and Lu, Hongtao},
  year = 2025,
  month = oct,
  pages = {25073--25082},
  publisher = {IEEE},
  address = {Honolulu, HI, USA},
  doi = {10.1109/ICCV51701.2025.02325},
  urldate = {2026-11-01},
  copyright = {https://doi.org/10.15223/policy-029},
  isbn = {979-8-3315-8775-8}
}

@misc{yang2026gs2ci,
  title = {{{GS}}{$^{2}$}{{CI}}: {{Robust Gaussian Splatting For Snapshot Compressive Imaging}} via {{Large Vision Model Priors}}},
  shorttitle = {{{GS}}{$^{2}$}{{CI}}},
  author = {Yang, Yanming and Song, Chenxi and Wang, Ping and Yuan, Xin and Zhang, Chi},
  year = 2026,
  month = aug,
  number = {arXiv:2608.13502},
  eprint = {2608.13502},
  primaryclass = {cs.CV},
  publisher = {arXiv},
  doi = {10.48550/arXiv.2608.13502},
  urldate = {2026-11-01},
  archiveprefix = {arXiv},
  howpublished = {arXiv preprint arXiv:2608.13502},
  langid = {english}
}

@article{yasuma2010generalized,
  title = {Generalized {{Assorted Pixel Camera}}: {{Postcapture Control}} of {{Resolution}}, {{Dynamic Range}}, and {{Spectrum}}},
  shorttitle = {Generalized {{Assorted Pixel Camera}}},
  author = {Yasuma, Fumihito and Mitsunaga, Tomoo and Iso, Daisuke and Nayar, Shree K},
  year = 2010,
  month = sep,
  journal = {IEEE Transactions on Image Processing},
  volume = {19},
  number = {9},
  pages = {2241--2253},
  issn = {1057-7149, 1941-0042},
  doi = {10.1109/TIP.2010.2046811},
  urldate = {2026-11-11},
  copyright = {https://ieeexplore.ieee.org/Xplorehelp/downloads/license-information/IEEE.html}
}

@inproceedings{yu2024mipsplatting,
  title = {Mip-{{Splatting}}: {{Alias-Free 3D Gaussian Splatting}}},
  shorttitle = {Mip-{{Splatting}}},
  booktitle = {2024 {{IEEE}}/{{CVF Conference}} on {{Computer Vision}} and {{Pattern Recognition}} ({{CVPR}})},
  author = {Yu, Zehao and Chen, Anpei and Huang, Binbin and Sattler, Torsten and Geiger, Andreas},
  year = 2024,
  month = jun,
  pages = {19447--19456},
  publisher = {IEEE},
  address = {Seattle, WA, USA},
  doi = {10.1109/CVPR52733.2024.01839},
  urldate = {2026-11-01},
  copyright = {https://doi.org/10.15223/policy-029},
  isbn = {979-8-3503-5300-6}
}

@inproceedings{yuan2016generalized,
  title = {Generalized Alternating Projection Based Total Variation Minimization for Compressive Sensing},
  booktitle = {2016 {{IEEE International Conference}} on {{Image Processing}} ({{ICIP}})},
  author = {Yuan, Xin},
  year = 2016,
  month = sep,
  pages = {2539--2543},
  publisher = {IEEE},
  address = {Phoenix, AZ, USA},
  doi = {10.1109/ICIP.2016.7532817},
  urldate = {2026-11-01},
  isbn = {978-1-4673-9961-6}
}

@incollection{zeng2024sahsci,
  title = {{{SAH-SCI}}: {{Self}}-Supervised {{Adapter}} for {{Efficient Hyperspectral Snapshot Compressive Imaging}}},
  shorttitle = {{{SAH-SCI}}},
  booktitle = {Computer {{Vision}} -- {{ECCV}} 2024},
  author = {Zeng, Haijin and Liu, Yuxi and Chen, Yongyong and Liu, Youfa and Peng, Chong and Su, Jingyong},
  editor = {Leonardis, Ale{\v s} and Ricci, Elisa and Roth, Stefan and Russakovsky, Olga and Sattler, Torsten and Varol, G{\"u}l},
  year = 2024,
  series = {Lecture {{Notes}} in {{Computer Science}}},
  volume = {15122},
  pages = {311--328},
  publisher = {Springer Nature Switzerland},
  address = {Cham},
  doi = {10.1007/978-3-031-73039-9_18},
  urldate = {2026-11-01},
  isbn = {978-3-031-73038-2 978-3-031-73039-9},
  langid = {english}
}

@inproceedings{zeng2026gaussian,
  title = {Gaussian {{Splatting}}-Based {{Low-Rank Tensor Representation}} for {{Multi-Dimensional Image Recovery}}},
  booktitle = {Proceedings of the {{IEEE}}/{{CVF Conference}} on {{Computer Vision}} and {{Pattern Recognition}} ({{CVPR}})},
  author = {Zeng, Yiming and Zhao, Xi-Le and Wu, Wei-Hao and Ji, Teng-Yu and Wang, Chao},
  year = 2026,
  pages = {19360--19369},
  urldate = {2026-11-01},
  langid = {english}
}

@inproceedings{zha2024r2gaussian,
  title = {{{R}}{$^2$}-{{Gaussian}}: {{Rectifying Radiative Gaussian Splatting}} for {{Tomographic Reconstruction}}},
  booktitle = {Advances in {{Neural Information Processing Systems}} ({{NeurIPS}})},
  author = {Zha, Ruyi and Lin, Tao Jun and Cai, Yuanhao and Cao, Jiwen and Zhang, Yanhao and Li, Hongdong},
  year = 2024,
  volume = {37},
  pages = {44907--44934},
  doi = {10.52202/079017-1427}
}

@inproceedings{zhang2019computational,
  title = {Computational {{Hyperspectral Imaging Based}} on {{Dimension-Discriminative Low-Rank Tensor Recovery}}},
  booktitle = {2019 {{IEEE}}/{{CVF International Conference}} on {{Computer Vision}} ({{ICCV}})},
  author = {Zhang, Shipeng and Wang, Lizhi and Fu, Ying and Zhong, Xiaoming and Huang, Hua},
  year = 2019,
  month = oct,
  pages = {10182--10191},
  publisher = {IEEE},
  address = {Seoul, Korea (South)},
  doi = {10.1109/ICCV.2019.01028},
  urldate = {2026-11-01},
  copyright = {https://doi.org/10.15223/policy-029},
  isbn = {978-1-7281-4803-8},
  langid = {english}
}

@inproceedings{zhang2024corgs,
  title = {{{CoR-GS}}: {{Sparse-View 3D Gaussian Splatting}} via {{Co}}-Regularization},
  booktitle = {Computer {{Vision}} -- {{ECCV}} 2024},
  author = {Zhang, Jiawei and Li, Jiahe and Yu, Xiaohan and Huang, Lei and Gu, Lin and Zheng, Jin and Bai, Xiao},
  year = 2024,
  series = {Lecture {{Notes}} in {{Computer Science}}},
  volume = {15059},
  pages = {335--352},
  publisher = {Springer Nature Switzerland},
  address = {Cham},
  doi = {10.1007/978-3-031-73232-4_19}
}

@inproceedings{zhang2024dual,
  title = {Dual {{Prior Unfolding}} for {{Snapshot Compressive Imaging}}},
  booktitle = {2024 {{IEEE}}/{{CVF Conference}} on {{Computer Vision}} and {{Pattern Recognition}} ({{CVPR}})},
  author = {Zhang, Jiancheng and Zeng, Haijin and Cao, Jiezhang and Chen, Yongyong and Yu, Dengxiu and Zhao, Yin-Ping},
  year = 2024,
  month = jun,
  pages = {25742--25752},
  publisher = {IEEE},
  address = {Seattle, WA, USA},
  doi = {10.1109/CVPR52733.2024.02432},
  urldate = {2026-11-01},
  copyright = {https://doi.org/10.15223/policy-029},
  isbn = {979-8-3503-5300-6},
  langid = {english}
}

@incollection{zhang2024gaussianimage,
  title = {{{GaussianImage}}: 1000 {{FPS Image Representation}} and {{Compression}} by {{2D Gaussian Splatting}}},
  shorttitle = {{{GaussianImage}}},
  booktitle = {Computer {{Vision}} -- {{ECCV}} 2024},
  author = {Zhang, Xinjie and Ge, Xingtong and Xu, Tongda and He, Dailan and Wang, Yan and Qin, Hongwei and Lu, Guo and Geng, Jing and Zhang, Jun},
  editor = {Leonardis, Ale{\v s} and Ricci, Elisa and Roth, Stefan and Russakovsky, Olga and Sattler, Torsten and Varol, G{\"u}l},
  year = 2024,
  series = {Lecture {{Notes}} in {{Computer Science}}},
  volume = {15067},
  pages = {327--345},
  publisher = {Springer Nature Switzerland},
  address = {Cham},
  doi = {10.1007/978-3-031-72673-6_18},
  urldate = {2026-11-01},
  isbn = {978-3-031-72672-9 978-3-031-72673-6},
  langid = {english}
}

@inproceedings{zhang2024improving,
  title = {Improving {{Spectral Snapshot Reconstruction}} with {{Spectral-Spatial Rectification}}},
  booktitle = {2024 {{IEEE}}/{{CVF Conference}} on {{Computer Vision}} and {{Pattern Recognition}} ({{CVPR}})},
  author = {Zhang, Jiancheng and Zeng, Haijin and Chen, Yongyong and Yu, Dengxiu and Zhao, Yin-Ping},
  year = 2024,
  month = jun,
  pages = {25817--25826},
  publisher = {IEEE},
  address = {Seattle, WA, USA},
  doi = {10.1109/CVPR52733.2024.02439},
  urldate = {2026-11-01},
  copyright = {https://doi.org/10.15223/policy-029},
  isbn = {979-8-3503-5300-6},
  langid = {english}
}

@inproceedings{zhang2025imagegs,
  title = {Image-{{GS}}: {{Content-Adaptive Image Representation}} via {{2D Gaussians}}},
  shorttitle = {Image-{{GS}}},
  booktitle = {{{SIGGRAPH Conference Papers}} '25},
  author = {Zhang, Yunxiang and Li, Bingxuan and Kuznetsov, Alexandr and Jindal, Akshay and Diolatzis, Stavros and Chen, Kenneth and Sochenov, Anton and Kaplanyan, Anton and Sun, Qi},
  year = 2025,
  month = aug,
  pages = {1--11},
  publisher = {ACM},
  address = {Vancouver, BC, Canada},
  doi = {10.1145/3721238.3730596},
  urldate = {2026-11-01},
  isbn = {979-8-4007-1540-2},
  langid = {english}
}

@article{zhang2026lrscgs,
  title = {Low-{{Rank Spectral Continuity Modeling With Gaussian Splatting}} for {{Hyperspectral Image Compression}}},
  author = {Zhang, Lili and Li, Yukun and Gao, Yang and Liu, Jiahui},
  year = 2026,
  journal = {IEEE Geoscience and Remote Sensing Letters},
  volume = {23},
  pages = {5507705},
  doi = {10.1109/LGRS.2026.3708431}
}

@misc{zhang2026voronoiguided,
  title = {Voronoi-Guided {{Bilateral 2D Gaussian Splatting}} for {{Arbitrary-Scale Hyperspectral Image Super-Resolution}}},
  author = {Zhang, Jie and You, Jinkun and Chen, Shi and Zhang, Yongshan and Zhou, Yicong},
  year = 2026,
  month = oct,
  number = {arXiv:2604.17727},
  eprint = {2604.17727},
  primaryclass = {cs.CV},
  publisher = {arXiv},
  doi = {10.48550/arXiv.2604.17727},
  urldate = {2026-11-01},
  archiveprefix = {arXiv},
  howpublished = {arXiv preprint arXiv:2604.17727}
}

@inproceedings{zhao2024untrained,
  title = {Untrained {{Neural Nets}} for {{Snapshot Compressive Imaging}}: {{Theory}} and {{Algorithms}}},
  shorttitle = {Untrained {{Neural Nets}} for {{Snapshot Compressive Imaging}}},
  booktitle = {Advances in {{Neural Information Processing Systems}}},
  author = {Zhao, Mengyu and Chen, Xi and Yuan, Xin and Jalali, Shirin},
  year = 2024,
  volume = {37},
  pages = {89752--89781},
  publisher = {Curran Associates, Inc.},
  doi = {10.52202/079017-2850},
  urldate = {2026-11-01}
}

@article{zheng2021deep,
  title = {Deep Plug-and-Play Priors for Spectral Snapshot Compressive Imaging},
  author = {Zheng, Siming and Liu, Yang and Meng, Ziyi and Qiao, Mu and Tong, Zhishen and Yang, Xiaoyu and Han, Shensheng and Yuan, Xin},
  year = 2021,
  month = feb,
  journal = {Photonics Research},
  volume = {9},
  number = {2},
  pages = {B18},
  issn = {2327-9125},
  doi = {10.1364/PRJ.411745},
  urldate = {2026-11-01},
  langid = {english}
}

@article{zhouwang2004image,
  title = {{{Image Quality Assessment: From Error Visibility to Structural Similarity}}},
  shorttitle = {Image Quality Assessment},
  author = {Wang, Zhou and Bovik, A.C. and Sheikh, H.R. and Simoncelli, E.P.},
  year = 2004,
  month = apr,
  journal = {IEEE Transactions on Image Processing},
  volume = {13},
  number = {4},
  pages = {600--612},
  issn = {1057-7149, 1941-0042},
  doi = {10.1109/TIP.2003.819861},
  urldate = {2026-11-11},
  copyright = {https://ieeexplore.ieee.org/Xplorehelp/downloads/license-information/IEEE.html}
}

@incollection{zhu2024fsgs,
  title = {{{FSGS}}: {{Real-Time Few-Shot View Synthesis Using Gaussian Splatting}}},
  shorttitle = {{{FSGS}}},
  booktitle = {Computer {{Vision}} -- {{ECCV}} 2024},
  author = {Zhu, Zehao and Fan, Zhiwen and Jiang, Yifan and Wang, Zhangyang},
  editor = {Leonardis, Ale{\v s} and Ricci, Elisa and Roth, Stefan and Russakovsky, Olga and Sattler, Torsten and Varol, G{\"u}l},
  year = 2024,
  series = {Lecture {{Notes}} in {{Computer Science}}},
  volume = {15097},
  pages = {145--163},
  publisher = {Springer Nature Switzerland},
  address = {Cham},
  doi = {10.1007/978-3-031-72933-1_9},
  urldate = {2026-11-01},
  isbn = {978-3-031-72932-4 978-3-031-72933-1},
  langid = {english}
}
